\documentclass [5p,hyperref]{elsarticle}

\usepackage[utf8]{inputenc}
\usepackage[figuresright]{rotating}
\usepackage{stfloats}
\usepackage{ragged2e}
\usepackage{amsmath}
\usepackage{amssymb}
\usepackage{booktabs}
\usepackage{makecell}
\usepackage{diagbox}
\usepackage{algorithm}
\usepackage{algorithmic}
\usepackage[flushleft]{threeparttable}
\usepackage{multirow}
\usepackage{url}
\usepackage{color}
\usepackage{tabularx} 
\usepackage[table,xcdraw]{xcolor}
\usepackage{float}

\usepackage{array}
\usepackage{makecell}
\usepackage{booktabs}

\newcolumntype{L}[1]{>{\raggedright\arraybackslash}m{#1}}
\newcolumntype{C}[1]{>{\centering\arraybackslash}m{#1}}

\usepackage{hyphenat}
\usepackage{microtype} 

\usepackage{array}  
\usepackage{graphicx}  
\usepackage{adjustbox}  
\usepackage{caption}
\usepackage{subcaption}
\makeatletter
\newcommand{\elsauthorbreak}{%
  \g@addto@macro\elsauthors{%
    \def\authorsep{\unskip,\space\protect\\}%
  }%
}
\makeatother

\usepackage{colortbl} 
\usepackage{lscape}
\usepackage[breaklinks=true,colorlinks=false]{hyperref}
 
\usepackage[pagewise]{lineno}
\usepackage{pdflscape}
\usepackage{footnote}
\definecolor{darkgray}{gray}{0.45} 
\hypersetup
{
	colorlinks=false,
	linkcolor=blue,
	filecolor=blue,
	urlcolor=blue,
	citecolor=cyan,
}
\makeatletter
\def\thefnote{\ifcase\c@fnote\or$\dagger$\or$\ddagger$\or\S\or\P\fi}
\makeatother

\journal{ISPRS Journal of Photogrammetry and Remote Sensing}

\begin{document}

\begin{frontmatter}

\title{Multimodal Large Language Models for Remote Sensing Image Understanding:  Domain-Specific or General-Purpose? }

\author[label1]{Qiwei Ma\fnref{equal}}
\ead{maqiwei@hnu.edu.cn}

\author[label2]{Chunping Qiu\fnref{equal}}

\author[label2]{Xinjun Cheng}
\elsauthorbreak
\author[label2]{Xiaoyu Zhang}
\author[label1]{Puhong Duan}
\author[label2]{Ke Yang\corref{cor1}}
\ead{yangke13@nudt.edu.cn}
\author[label1,label3]{Xudong Kang\corref{cor1}}
\author[label1]{Shutao Li}

\fntext[equal]{These authors contributed equally to this work.}
\cortext[cor1]{Corresponding author}

\affiliation[label1]{organization={School of Artificial Intelligence and Robotics, Hunan University},
      city={Changsha},
      postcode={410082}, 
      country={China}}

\affiliation[label2]{organization={Intelligent Game and Decision Lab (IGDL)},
      city={Beijing},
      postcode={100091}, 
      country={China}}

\affiliation[label3]{organization={Yuelushan Center for Industrial Innovation},
      city={Changsha},
      postcode=410082, 
      country={China}}

\begin{abstract}

The rapid development of multimodal large language models (MLLMs) has introduced a flexible paradigm for remote sensing image scene understanding (RSISU), enabling natural-language interaction with remote sensing imagery. However, a systematic understanding of the capability boundaries, cross-task generalization, and task-specific limitations of existing remote sensing MLLMs (RS-MLLMs) is still lacking. This paper presents a systematic survey and diagnostic evaluation of MLLMs for RSISU. We review the technical evolution of RS-MLLMs, focusing on model design, multimodal learning, training data, and downstream capabilities. We further compare RS-MLLMs with general-purpose computer vision MLLMs (CV-MLLMs) across diverse RSISU tasks and benchmarks. RS-MLLMs remain competitive in domain-specific settings, particularly remote sensing visual grounding and high-resolution visual question answering. More notably, general-purpose CV-MLLMs can match or even outperform these specialized models on several RSISU tasks without remote sensing-specific fine-tuning. These findings demonstrate the strong transferability of general-purpose CV-MLLMs and show that current RS-MLLMs do not consistently outperform them across diverse RSISU tasks. Current MLLMs also face limitations in spatial and relational reasoning, fine-grained visual understanding, instruction diversity, and generalization across heterogeneous task formats. Based on these findings, we outline future directions toward reliable evaluation, multimodal and high-resolution reasoning, efficient deployment, and tool-augmented remote sensing agents. This survey provides a systematic reference for developing robust, generalizable, and practical MLLMs for RSISU.

\end{abstract}


\end{frontmatter}

\switchlinenumbers

\section{Introduction}
\label{sec:introduction}

Remote sensing image scene understanding (RSISU) aims to interpret complex earth observation scenes beyond image-level recognition. It covers scene classification (SC), image captioning (IC), visual question answering (VQA), and visual grounding (VG). It also involves object localization, attribute analysis, spatial reasoning, and change understanding~\citep{Williams2023StructuredGM,HUANG2024103939,ZHAO2024103672}. Remote sensing (RS) images differ substantially from natural images. They are characterized by overhead viewpoints, large variations in object scale, dense object distributions, and diverse sensor modalities. Their interpretation also depends strongly on spatial resolution, viewing geometry, illumination conditions, and geographic context. Models must recognize individual objects and land-cover categories while understanding their spatial relationships, scene-level semantics, and geographic context. Although recent foundation models (FMs) and multimodal large language models (MLLMs) have significantly advanced general visual understanding, these RS-specific challenges remain difficult to address.

Foundation models are large-scale pre-trained models that can be adapted to downstream tasks through fine-tuning, few-shot learning, or zero-shot transfer. Representative examples include language FMs such as GPT-3~\citep{GPT3}, vision FMs such as MAE~\citep{MAE}, and vision-language FMs such as CLIP~\citep{CLIP} and BLIP~\citep{Li2022BLIPBL}. Early RS FMs mainly focused on visible optical imagery. Models such as SatMAE~\citep{Cong2022SatMAEPT}, ScaleMAE~\citep{Reed2022ScaleMAEAS}, RingMo~\citep{Sun2023RingMoAR}, and GFM~\citep{Mendieta2023GFMBG} learned transferable representations from large-scale RS image collections. More recent RS FMs have expanded beyond optical imagery to hyperspectral, synthetic aperture radar (SAR), and multimodal observations. SpectralGPT~\citep{spectralgpt} and HyperSIGMA~\citep{hypersigma} focus on hyperspectral representation learning, whereas SARJEPA~\citep{SARJEPA} and SARMAE~\citep{sarmae} target SAR imagery. SkySense~\citep{skysense} and TerraFM~\citep{terrafm} further explore unified representations across multiple RS modalities. These models provide important visual and cross-modal backbones for the development of RS-oriented MLLMs.

MLLMs extend conventional large language models (LLMs) with multimodal perception and cross-modal reasoning capabilities. They support instruction-following tasks such as VQA, image captioning, visual grounding, image-based dialogue, and multimodal reasoning~\citep{Wu2023Survey}. A typical MLLM consists of a visual encoder, a modality alignment module, and an LLM. The visual encoder extracts image features, while the alignment module maps these features into representations that can be processed by the LLM. This design enables more flexible interaction than conventional task-specific vision models. Proprietary systems such as GPT-4V and Gemini have demonstrated strong multimodal reasoning capabilities. Open-source models such as mPLUG-Owl have further demonstrated the potential of adapting pre-trained knowledge to multimodal tasks at reduced training cost~\citep{ye2024mplugowlmodularizationempowerslarge}. These advances create new opportunities for RSISU, which requires the joint modeling of visual perception, geographic semantics, and natural-language interaction.

\begin{figure*}[!t]
	\centering
	\includegraphics[width=\textwidth]{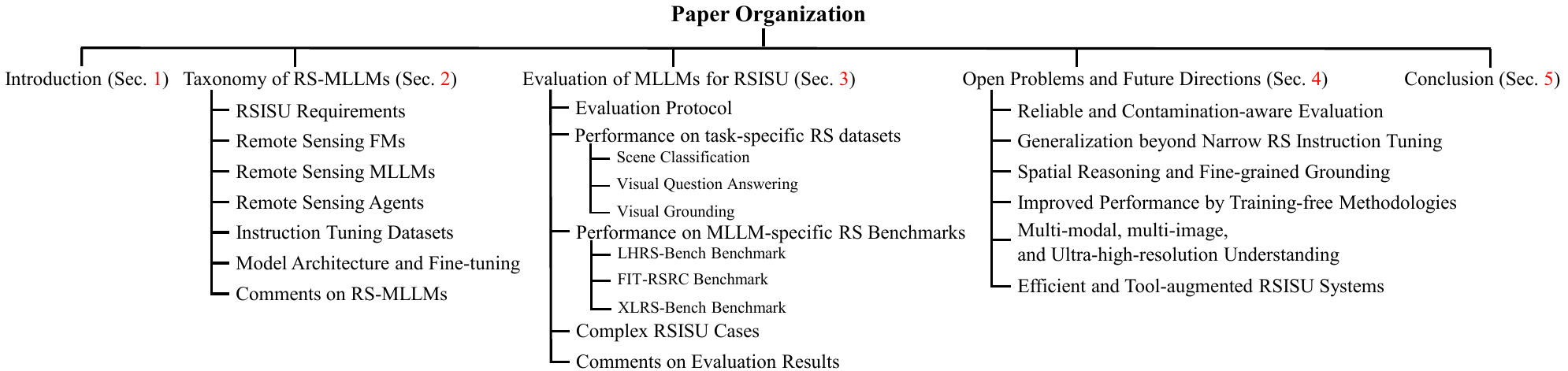}
	\caption{Diagram of paper organization.
	}
	\label{fig:organization}
\end{figure*}

Early studies explored the direct application of general-purpose MLLMs and tool-augmented systems to RS and geospatial tasks. \citet{rs15133232} evaluated Visual ChatGPT on scene classification, edge detection, line detection, and segmentation. The model showed potential in relatively simple scenes but performed poorly in complex urban environments. Similarly, \citet{Mai2023OnTO} examined several language, vision, and multimodal foundation models for geospatial artificial intelligence. These models performed well on text-based geospatial tasks under zero-shot and few-shot settings. However, they remained limited when tasks required the integration of multiple data modalities~\citep{Qiu2023}. \cite{zhang2024good} further evaluated GPT-4V on Earth observation tasks. GPT-4V achieved promising results in aerial landmark recognition and image captioning but struggled with object localization, counting, spatial reasoning, and change detection. These early findings revealed several limitations in applying general-purpose MLLMs to RSISU and motivated the development of RS-specific models. However, general-purpose MLLMs have since advanced rapidly in visual perception and multimodal reasoning. This progress raises a central question: \textbf{Do RS-specific MLLMs consistently outperform general-purpose models across diverse RSISU tasks?}

In this survey, MLLMs primarily developed for general visual understanding are referred to as CV-MLLMs. Models specifically designed or adapted for RS tasks are referred to as RS-MLLMs. Early RS-MLLMs mainly focused on image captioning and VQA. More recent models have expanded toward multitask dialogue, visual grounding, spatial reasoning, and large-scale scene understanding~\citep{zhou2024visionlanguagegeofoundationmodelsurvey}. Representative models include RSGPT~\citep{Hu2023RSGPTAR}, GeoChat~\citep{GeoChat}, SkyEyeGPT~\citep{Zhan2024SkyEyeGPTUR}, EarthGPT~\citep{EarthGPT}, LHRS-Bot~\citep{LHRS-Bot}, H\textsuperscript{2}RSVLM~\citep{pang2025vhm}, and SkySenseGPT~\citep{SkySenseGPT}. These studies have contributed to RS instruction tuning, multimodal dialogue, grounded understanding, dataset construction, and complex scene interpretation. In parallel, increasing attention has been given to RS-oriented vision encoders and vision-language foundation models. These components are important for handling the sensor diversity, scale variation, and geographic semantics of RS imagery~\citep{skyscript,GRAFT}.

\begin{figure*}[!t]
	\centering
	\includegraphics[width=\textwidth]{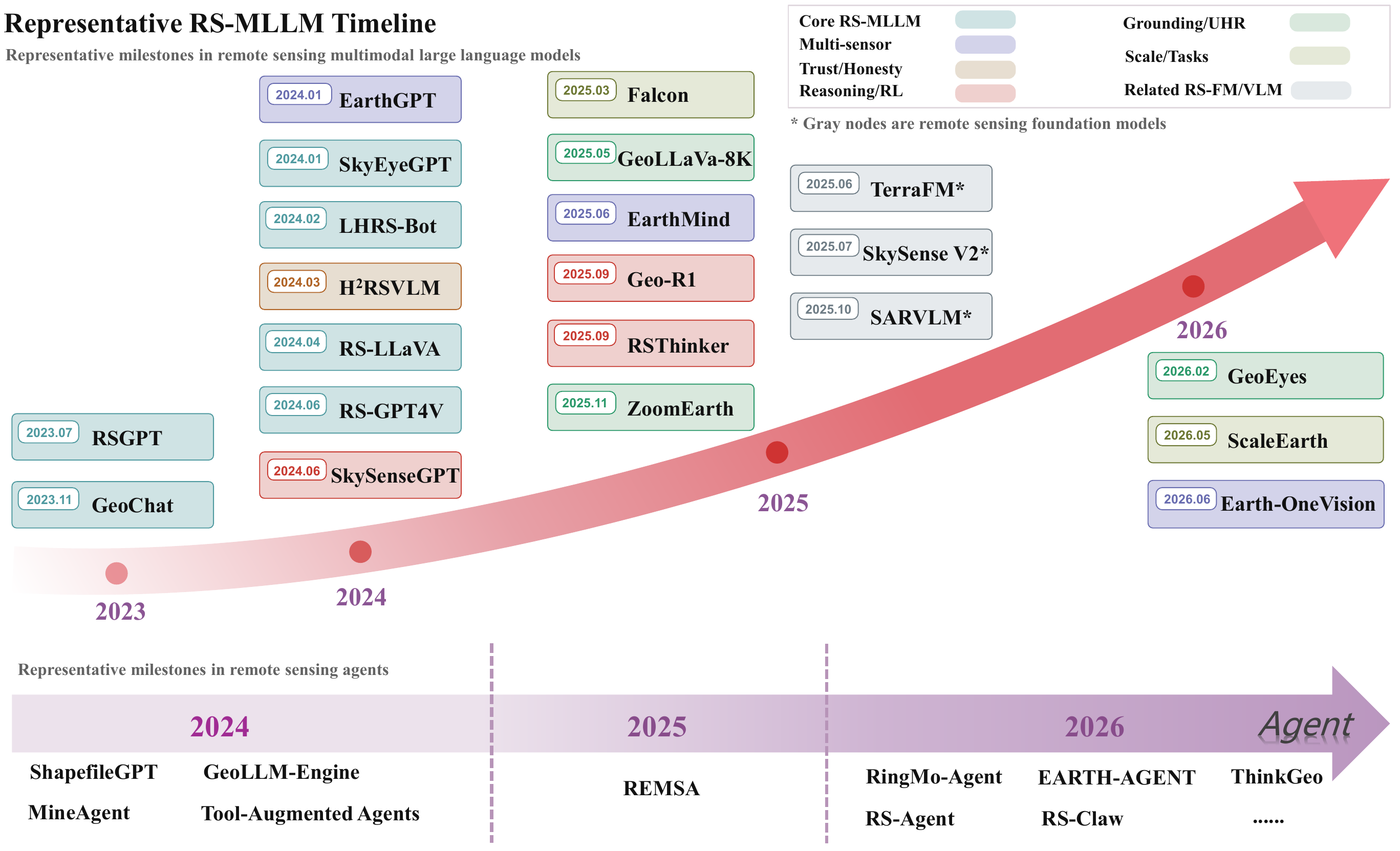}
	\caption{Representative timeline of remote-sensing multimodal large language models (RS-MLLMs) and related works. 
    Different colors denote the dominant technical focus of each work: \textbf{Core RS-MLLM} refers to general-purpose RS multimodal models covering captioning, VQA, chat, and instruction tuning; \textbf{Multi-sensor} highlights models designed for multi-sensor or cross-modal RS inputs, such as optical, SAR, infrared, and multi-source Earth observation data; \textbf{Trust/Honesty} denotes works focusing on reliability, honesty, refusal ability, and hallucination mitigation; \textbf{Reasoning/RL} represents methods emphasizing geospatial reasoning, chain-of-thought reasoning, or reinforcement learning-based tuning such as GRPO; \textbf{Grounding/UHR} indicates models targeting visual grounding, evidence grounding, ultra-high-resolution imagery, and active crop/zoom perception; \textbf{Scale/Tasks} denotes works emphasizing large-scale data, broad task coverage, or scale-aware modeling. 
    Gray nodes indicate related RS foundation models or vision-language models that provide contextual background but are not treated as the main instruction-following RS-MLLM line. 
    The lower axis summarizes the development of AI-agent research in the remote-sensing domain.
	}
	\label{fig:RS-MLLMs}
\end{figure*}

Recent RS-MLLMs have further advanced toward finer-grained, sensor-aware, and reasoning-oriented understanding. Falcon~\citep{yao2025falcon} constructs a large-scale multitask instruction-tuning corpus to support image-, region-, and pixel-level interpretation. GeoLLaVA-8K~\citep{wang2026geollava} and ZoomEarth~\citep{liu2026zoomearth} focus on ultra-high-resolution RS imagery. They employ high-resolution encoding, adaptive cropping, or active perception to process large scenes. EarthGPT-X~\citep{EarthGPT-X} extends RS-MLLMs to multi-source spatial understanding by integrating optical, SAR, and infrared imagery with visual prompting. Geo-R1~\citep{zhang2026geo} introduces reinforcement fine-tuning for few-shot geospatial referring expression understanding. RemoteShield~\citep{remoteshield} investigates robustness against realistic visual and textual perturbations. Collectively, these advances show that RS-MLLM research is moving beyond conventional captioning and VQA. Emerging priorities include high-resolution perception, spatial grounding, multi-source interpretation, reasoning-oriented post-training, and reliable deployment.

Despite this progress, existing surveys have not yet provided a systematic and diagnostic understanding of RS-MLLMs for RSISU. Most reviews focus on RS foundation models, vision-language datasets, or general MLLM architectures~\citep{li2024vision}. The effects of RS-specific instruction tuning, RS-oriented visual encoders, and multimodal alignment remain insufficiently examined. Existing models are also evaluated using different datasets, prompts, metrics, and inference protocols. These differences make direct comparisons difficult. More importantly, it remains unclear whether RS-specific instruction tuning provides consistent advantages over recent general-purpose MLLMs. This issue is increasingly important because CV-MLLMs have rapidly improved in visual perception and cross-modal reasoning. In contrast, many RS-MLLMs are trained with relatively limited instruction data and evaluated under heterogeneous protocols.

As summarized in Table~\ref{tab:comparison_surveys}, this survey differs from previous reviews by combining a systematic taxonomy with a diagnostic evaluation. We examine the technical development of RS-MLLMs and analyze the roles of model architecture, visual encoders, multimodal alignment, instruction tuning, and training data. We further compare RS-MLLMs with recent open-source and proprietary CV-MLLMs under representative RSISU settings. Our analysis investigates whether domain-specific adaptation leads to consistent performance gains. The results reveal that the relative strengths of RS-MLLMs and CV-MLLMs vary considerably across tasks. RS-MLLMs remain competitive in several domain-specific settings. More notably, general-purpose CV-MLLMs can match or even outperform specialized RS-MLLMs on several RSISU tasks without remote sensing-specific fine-tuning. This diagnostic perspective provides an empirical basis for future model development, benchmark construction, and evaluation protocol design. We focus primarily on end-to-end and instruction-tuned RS-MLLMs, while discussing RS agents as an important direction toward executable and verifiable geospatial intelligence.

We searched Google Scholar, Web of Science, IEEE Xplore, Scopus, Elsevier, Springer, and arXiv for relevant studies published through July 2026. The search queries included ``remote sensing multimodal large language model,'' ``remote sensing vision--language model,'' ``Earth observation multimodal model,'' ``remote sensing agent,'' and combinations of these terms. We included studies that developed, adapted, or systematically evaluated instruction-following multimodal models for remote sensing image understanding. Studies limited to unimodal representation learning, conventional image classification, or language-only geospatial tasks were excluded unless they provided essential architectural, methodological, or historical context for the development of RS-MLLMs.

Our main findings and contributions are summarized as follows:
\begin{itemize}

\item We provide a systematic review of RS-MLLMs for RSISU. The review covers model architectures, key components, instruction-tuning data, training strategies, and evaluation settings within the broader development of visual and vision-language foundation models.

\item We systematically compare state-of-the-art RS-MLLMs with open-source and proprietary CV-MLLMs across representative RSISU datasets and tasks. This comparison provides a diagnostic assessment beyond the results reported by individual studies.

\item We show that the relative performance of RS-MLLMs and CV-MLLMs is strongly task-dependent. General-purpose CV-MLLMs can match or outperform specialized RS-MLLMs on several RSISU tasks without RS-specific fine-tuning. We also identify limitations related to instruction diversity, evaluation protocols, and potential benchmark contamination.

\item We discuss future directions for reliable evaluation, spatially grounded reasoning, training-free adaptation, multimodal and multitemporal understanding, high-resolution image processing, efficient deployment, and RS agents.

\end{itemize}

The remainder of this survey is organized as follows. Section~\ref{sec:taxonomy} presents the taxonomy of RS-MLLMs and reviews their instruction-tuning datasets, model architectures, training strategies, and key characteristics. Section~\ref{sec:performance} compares RS-MLLMs and CV-MLLMs across representative RSISU tasks. Section~\ref{sec:challenge} discusses the main challenges and future research directions. Finally, Section~\ref{sec:conclusion} concludes the survey.

\begin{table*}[!t]
\centering
\scriptsize
\setlength{\tabcolsep}{3pt}
\renewcommand{\arraystretch}{1.2}

\caption{Comparison between this survey and representative recent surveys related to remote sensing foundation models, vision-language models, and multimodal large language models.}
\label{tab:comparison_surveys}

    \begin{tabular}{L{1.7cm}L{2.35cm}C{1.85cm}C{2.05cm}L{1.95cm}L{6.95cm}}
    \toprule
    \textbf{Survey} 
    & \textbf{Main Scope} 
    & \textbf{RS-MLLM Taxonomy} 
    & \textbf{Instruction-tuning and Datasets} 
    & \textbf{Diagnostic Evaluation} 
    & \textbf{Main Difference from this Survey} \\
    \midrule
    
    Li et al. \citep{li2024vision}
    & Vision-language models in remote sensing
    & Partial
    & Limited
    & No unified RS-MLLM vs. CV-MLLM diagnosis
    & Focuses on RS vision-language tasks, while RS-MLLM architectures, instruction tuning, and agentic/UHR models are less emphasized. \\
    
    Zhou et al. \citep{zhou2024visionlanguagegeofoundationmodelsurvey}
    & Vision-language geo-foundation models
    & Partial
    & Partial
    & No systematic diagnostic evaluation
    & Reviews VLGFMs from data, architecture, and geospatial task perspectives, without unified RS-MLLM vs. CV-MLLM capability diagnosis. \\
    
    Xiao et al. \citep{Xiao2025RSFMSurvey}
    & Foundation models for remote sensing and Earth observation
    & Partial
    & Limited
    & Broad benchmark comparison
    & Provides a broad RS foundation-model survey, but is less focused on instruction-following RS-MLLMs and task-specific generalization. \\
    
    Huo et al. \citep{huo2025remote}
    & General and domain-specific foundation models for remote sensing
    & Limited
    & Limited
    & Limited
    & Emphasizes foundation-model paradigms rather than RS-MLLM-specific instruction following, reasoning, and diagnostic evaluation. \\
    
    Yang et al. \citep{yang2025survey}
    & Multimodal geospatial foundation models
    & Partial
    & Partial
    & Broad downstream evaluation
    & Takes a modality-driven view of multimodal GFMs, while our survey centers on instruction-following RS-MLLMs for RS image understanding. \\
    
    Zhou et al. \citep{zhou2025advances}
    & Multimodal RS foundation models for EO downstream tasks
    & Partial
    & Partial
    & General discussion
    & Develops a vision--X taxonomy, but gives less attention to RS-MLLM vs. CV-MLLM comparison, contamination-aware evaluation, and failure modes. \\
    
    Xu et al. \citep{Xu2025EOMLLMSurvey}
    & Earth observation multimodal large language models
    & Extensive
    & Extensive
    & Limited
    & Reviews EO-MLLM frameworks and technologies, but lacks broad empirical comparison with state-of-the-art CV-MLLMs on RSISU benchmarks. \\
    
    \textbf{Ours}
    & \textbf{RS-MLLMs for remote sensing image understanding}
    & \textbf{\makecell[c]{Compre-\\hensive}}
    & \textbf{\makecell[c]{Compre-\\hensive}}
    & \textbf{Yes}
    & \textbf{Provides a comprehensive RS-MLLM taxonomy and benchmark-level RS-MLLM vs. CV-MLLM diagnosis, covering architectures, encoders, instruction tuning, datasets, modalities, task capabilities, contamination-aware evaluation, spatial/UHR reasoning, multi-image/multi-temporal reasoning, training-free adaptation, and RS agents.} \\

\bottomrule
\end{tabular}
\end{table*}

\begin{table*}[t]  
  \centering  
  \scriptsize
  \caption{RS specific visual encoders within visual FMs and visual-language FMs.} 
  \renewcommand{\arraystretch}{1.2}
  \begin{tabularx}{\linewidth}{l c X c l}  
    \toprule  
    Method & Modality & Visual Encoder & Text Encoder & \# images or i-t pairs\\ \midrule 
    Geograph \citep{Ayush2020GeographyAwareSL}   &       &   ResNet50  &       & 0.9M \\
    RSP \citep{Wang2022AnES}                     & \multirow{8}[0]{*}{Visible} & Swin-Tiny, ViTAEv2-S & \multirow{17}[0]{*}{-} & 1M \\
    RVSA \citep{Wang2022AdvancingPV}             &          & ViTAE-B &       & 1M  \\
    SatMAE \citep{Cong2022SatMAEPT}              &          & ViT-L &       & 0.7M \\
    Scale MAE  \citep{Reed2022ScaleMAEAS}        &          & ViT-L &       & 0.4M \\
    RingMo \citep{Sun2023RingMoAR}               &          & Swin-T Base &       &  1M \\
    Billion  \citep{Cha2023ABF}                  &          & ViT G12 &       & 1M \\
    GFM  \citep{Mendieta2023GFMBG}               &          & Swin  &       & 0.6M \\
    SpectralGPT~\citep{spectralgpt}              &  \multirow{3}[1]{*}{Spectral} & ViT-B &       & 1M\\
    HyperSIGMA~\citep{hypersigma}                &          & ViT-B, ViT-L, ViT-H &       & 0.44M\\
    HyperFree~\citep{hyperfree}                  &          & ViT-B &       & 0.15M\\
    SARJEPA~\citep{SARJEPA}                      & \multirow{3}[1]{*}{SAR}      & ViT-B &       & 0.095M\\
    SARATRX~\citep{SARATRX}                      &          & HiViT-B &       & 0.18M\\
    SARMAE~\citep{sarmae}                        &          & ViT-B,ViT-L &       & 1.3M \\
    SkySense~\citep{skysense}                    & \multirow{3}[1]{*}{Multimodal}      & Swin-H, ViT-L &       & 21.5M \\
    TerraFM~\citep{terrafm}                      &          & ViT-B/16, ViT-L/16 &       & 18.7M\\
    SkySense V2~\citep{skysensev2}               &          & Unified Transformer &       & 21M\\
    \midrule  
    SkyCLIP~\citep{skyscript}                    & \multirow{9}[1]{*}{VL} & ViT-L & \multirow{9}[1]{*}{Transformer} & 2.6M \\
    RemoteCLIP~\citep{Liu2023RemoteCLIPAV}       &          & ResNet-50, ViT-B/3, ViT-L/14 &       & 828K\\
    GRAFT~\citep{GRAFT}                 &          & ViT-B/32, ViT-B/16 &       & 2M        \\
    GeoRSCLIP~\citep{Zhang2023RS5MAG}            &          & ViT-B/32, ViT-L/14    &       & 5M     \\ 
    HQRS~\citep{HQRS}                            &          & ViT-B/32, ViT-L/14    &       & 1.26M     \\ 
    SAR-TEXT~\citep{SARTEXT}                      &          & ViT-L/14          &       & 130K     \\ 
    SARVLM~\citep{sarvlm}                        &          & ResNet-50, ViT-B/32, ViT-B/16, ViT-L/14 &       & 1.39M  \\
    SARCLIP~\citep{sarclip-isprs}                &          & \makecell[l]{ResNet-50, ResNet-101, \\ ViT-B/32, ViT-B/16, ViT-L/14} & & 0.43M  \\
    HyperCap~\citep{HyperCap}                    &          & \makecell[l]{DBCTNet, FAHM, \\ 3D-ConvSST, 3D-RCNet} &   & 21.37K    \\
    \bottomrule
  \end{tabularx}  
  \label{tab:rsEncoder}
\end{table*}

\begin{figure*}[t]
	\centering
	\includegraphics[width=\textwidth]{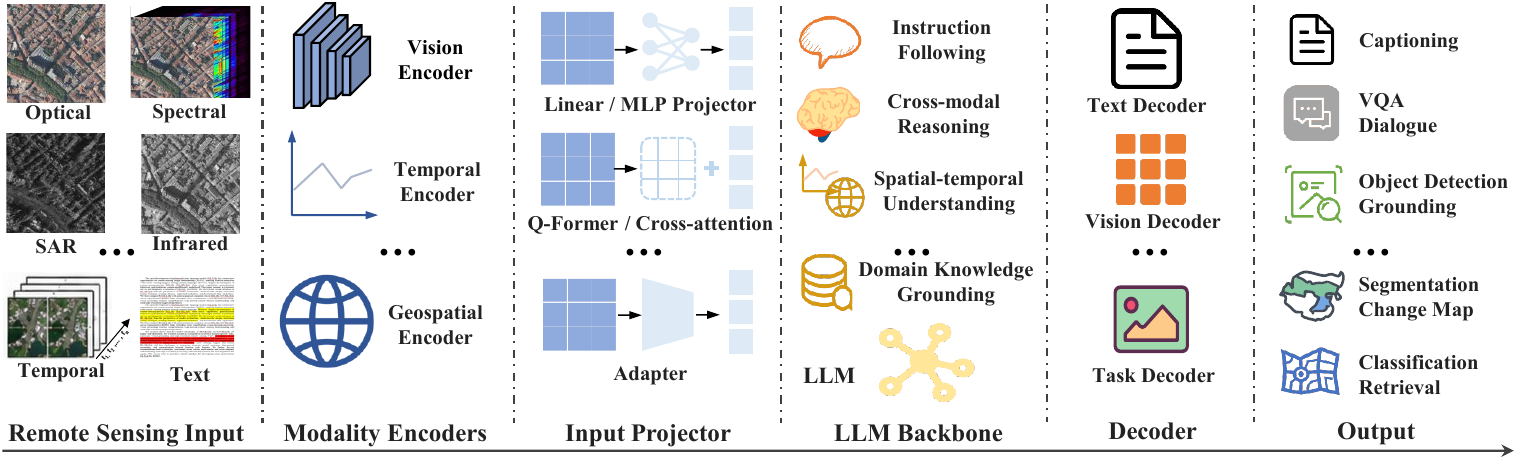}
	\caption{General model architecture of MLLMs.}
	\label{fig:modelarchitecture}
\end{figure*}
\section{Taxonomy of RS-MLLMs}
\label{sec:taxonomy}
This section will outline these RSISU-specific challenges and subsequently review the current landscape of RS-FMs, RS-MLLMs and RS-Agents. Following this discussion, a detailed analysis of the model architectures and datasets supporting existing RS-MLLM research will be presented.

\begin{table*}[!t] 
\scriptsize
\caption{Specific model architecture and tuning details of SOTA RS-MLLMs (\textcolor[rgb]{0, .69, .941}{${*}$}: frozen, \textcolor[rgb]{.439, .678, .278}{$\triangle$}: fine-tuning with parameter-efficient techniques, \textcolor[rgb]{1, 0, 0}{$\lozenge$}: fine-tuning, black text without symbols indicates different strategies applied at various stages.)}
\begin{tabular}{  
    >{\centering\arraybackslash}m{2.5cm} 
    >{\centering\arraybackslash}m{2cm}
    >{\centering\arraybackslash}m{2cm}  
    >{\centering\arraybackslash}m{1.5cm} 
    >{\centering\arraybackslash}m{1cm}  
    >{\raggedright\arraybackslash}m{7cm}  
}   
    \toprule  
    \textbf{Model} & \textbf{Visual Encoder} & \textbf{Input Projector} & \textbf{LLM} & \textbf{Hardware} & \textbf{Training/Tuning Details} \\
    \midrule  
    \makecell{RSGPT \\ \citep{Hu2023RSGPTAR}} & \textcolor[rgb]{0, .69, .941}{ViT-G (EVA)$^{*}$} & \textcolor[rgb]{1, 0, 0}{a Q-Former + a linear layer\raisebox{1ex}{\footnotesize$\lozenge$}} & \textcolor[rgb]{0, .69, .941}{Vicuna-13B$^{*}$} & 8*A100 (80G) & initialization with InstructBLIP; fine-tuning I-P using RSICap dataset \\
    \midrule  
    
    \makecell{SkyEyeGPT \\ \citep{Zhan2024SkyEyeGPTUR}} & \textcolor[rgb]{0, .69, .941}{ViT-G (EVA-CLIP)$^{*}$} & \textcolor[rgb]{.439, .678, .278}{a linear layer\raisebox{1ex}{\footnotesize$\triangle$}} & \textcolor[rgb]{.439, .678, .278}{LLaMA2-7B\raisebox{1ex}{\footnotesize$\triangle$}} & 4*3090Ti (24G) & initialization with MiniGPT-v2; LoRA fine-tuning I-P and LLM; stage 1: 35 epochs on single-task instructions; stage 2: 5 epochs on multi-round multitask conversation with reduced single-task instruction ratio \\
    \midrule  
    
    \makecell{EarthGPT \\ \citep{EarthGPT}} & \textcolor[rgb]{0, .69, .941}{ViT-L (DINOv2)$^{*}$ + ConvNeXt-L (CLIP)$^{*}$} & \textcolor[rgb]{0, .69, .941}{a linear layer$^{*}$} & \textcolor[rgb]{1, 0, 0}{LLaMA2-7B\raisebox{1ex}{\footnotesize$\lozenge$}} & 16*A100 (80G) & cross-modal mutual comprehension stage: training on LAION-400M, COCO Caption from scratch; unified multitask tuning phase: training LLM on MMRS-1M and randomly initializing I-P \\
    \midrule  
    
    \makecell{LHRS-Bot \\ \citep{LHRS-Bot}} & \textcolor[rgb]{0, .69, .941}{ViT-L (CLIP)$^{*}$} & \textcolor[rgb]{1, 0, 0}{a Vision Perceiver (six layers of cross-attention and MLP)\raisebox{1ex}{\footnotesize$\lozenge$}} & \textcolor[rgb]{.439, .678, .278}{LLaMA2-7B\raisebox{1ex}{\footnotesize$\triangle$}} & 8*V100 (32G) & stage 1: Pretraining  of the input projector (I-P) with the LHRS-Align dataset; stage 2 involves training the I-P and applying LoRA fine-tuning to the LLM using multitask instruction datasets and 4K detailed description data from the LHRS-Instruct dataset. Stage 3 consists of supervised fine-tuning with the LHRS-Instruct dataset, 20K selected data from the LLaVA complex reasoning dataset, and multitask datasets from stage 2, utilizing a lower sampling ratio. \\
    \midrule  
    
    \makecell{GeoChat\\ \citep{GeoChat}} & \textcolor[rgb]{0, .69, .941}{ViT-L (CLIP)$^{*}$} & \textcolor[rgb]{0, .69, .941}{a two-layer MLP$^{*}$} & \textcolor[rgb]{.439, .678, .278}{Vicuna-v1.5-7B\raisebox{1ex}{\footnotesize$\triangle$}} & 3*A100 (40G) & initialization with LLaVA-v1.5; LoRA fine-tuning LLM; two stages of training: with all datasets for 1 epoch (2400 steps), then with grounding dataset for an additional 1600 steps \\
    \cmidrule{1-6}  
    
    \makecell{H\textsuperscript{2}RSVLM \\ \citep{pang2025vhm}}   & ViT-L (CLIP) & \textcolor[rgb]{1, 0, 0}{{a two-layer MLP}\raisebox{1ex}{\footnotesize$\lozenge$}} & \textcolor[rgb]{1, 0, 0}{Vicuna-v1.5-7B (13B)\raisebox{1ex}{\footnotesize$\lozenge$}} & 16*A100 (80G) & initialization with LLaVA-v1.5; pretraining stage: fine-tuning V-E, LLM, and I-P on HqDC-1.4M dataset, 1 epoch, 5400 steps; SFT LLM and I-P on HqDC-Instruct, RSSA, RS-Specialized-Instruct, RS-ClsQaGrd-Instruct \\
    \cmidrule{1-6}  
    
    \makecell{RS-LLaVA \\ \citep{RS-LLaVA}} & \textcolor[rgb]{0, .69, .941}{ViT-L (CLIP)$^{*}$} & \textcolor[rgb]{.439, .678, .278}{a two-layer MLP\raisebox{1ex}{\footnotesize$\triangle$}} & \textcolor[rgb]{.439, .678, .278}{Vicuna-v1.5-13B\raisebox{1ex}{\footnotesize$\triangle$}} & 2*A6000 (48G) & initialization with LLaVA-v1.5; pretraining I-P using general text–image pairs; LoRA fine-tuning LLM \\
    \cmidrule{1-6}  
    
    \makecell{SkySenseGPT \\ \citep{SkySenseGPT}} & \textcolor[rgb]{0, .69, .941}{ViT-L (CLIP)$^{*}$} & \textcolor[rgb]{1, 0, 0}{a two-layer MLP\raisebox{1ex}{\footnotesize$\lozenge$}} & \textcolor[rgb]{.439, .678, .278}{Vicuna-v1.5-7B\raisebox{1ex}{\footnotesize$\triangle$}} & 4*A100 (40G) & initialization with LLaVA-v1.5; fine-tuning I-P, and LoRA fine-tuning LLM using training and validation set of FIT-RS and additional publication sets \\
    \cmidrule{1-6}  
    
    \makecell{RS-GPT4V \\ \citep{rsgpt}} & \textcolor[rgb]{0, .69, .941}{ViT-L (CLIP)$^{*}$} & a two-layer MLP & Vicuna-v1.5-7B & 4*A800 (80G) & initialization with LLaVA-v1.5; three kinds of tuning with a total of 14,666 training steps, corresponding to 1 epoch: Full-FT, LoRA fine-tuning, and MoE-LoRA fine-tuning \\
    \midrule

    \makecell[c]{Falcon\\ \citep{yao2025falcon}}  & \textcolor[rgb]{1, 0, 0}{an image encoder\raisebox{1ex}{\footnotesize$\lozenge$}} 
    & -- 
    & \textcolor[rgb]{1, 0, 0}{transformer-based encoder-decoder, 0.7B\raisebox{1ex}{\footnotesize$\lozenge$}} 
    & 160*A100 
    & initialization with pretrained weights from Florence-2; supervised fine-tuning on Falcon SFT with about 78M samples and 5.6M remote sensing images across 14 tasks; image size 448$\times$448, output length 4096, batch size 640, trained for 4 days \\
    \midrule

    \makecell[c]{Geo-R1\\\citep{zhang2026geo}} & \textcolor[rgb]{1, 0, 0}{Qwen2.5VL visual encoder\raisebox{1ex}{\footnotesize$\lozenge$}} 
    & \textcolor[rgb]{1, 0, 0}{Qwen2.5VL built-in projector\raisebox{1ex}{\footnotesize$\lozenge$}} 
    & \textcolor[rgb]{1, 0, 0}{Qwen2.5VL-3B-Instruct\raisebox{1ex}{\footnotesize$\lozenge$}} 
    & 8*H100 
    & initialization with Qwen2.5VL-3B-Instruct; reinforcement fine-tuning with GRPO on VRSBench-FS, NWPU-FS and EarthReason-FS; trained for 30 epochs with early stopping; thinking prompts are used for RL-based paradigms \\
    \midrule

    \makecell{ZoomEarth\\ \citep{liu2026zoomearth}} & \textcolor[rgb]{1, 0, 0}{Qwen2.5VL visual encoder\raisebox{1ex}{\footnotesize$\lozenge$}} 
    & \textcolor[rgb]{1, 0, 0}{Qwen2.5VL built-in projector\raisebox{1ex}{\footnotesize$\lozenge$}} 
    & \textcolor[rgb]{1, 0, 0}{Qwen2.5VL-3B\raisebox{1ex}{\footnotesize$\lozenge$}} 
    & 8*A800 
    & initialization with Qwen2.5VL-3B; stage 1: SFT on LRS-GRO to learn RS knowledge and tool-invocation format; stage 2: GRPO training with IoU reward, Region-Guided reward, answer reward and format reward; adaptive cropping-zooming tool is used for active perception \\
    \midrule

    \makecell{GeoLLaVA-8K\\ \citep{wang2026geollava}}  & \textcolor[rgb]{1, 0, 0}{LLaVA-Next visual encoder\raisebox{1ex}{\footnotesize$\lozenge$}} 
    & \textcolor[rgb]{1, 0, 0}{projection layer\raisebox{1ex}{\footnotesize$\lozenge$}}
    & \textcolor[rgb]{1, 0, 0}{LLaVA-Next-7B\raisebox{1ex}{\footnotesize$\lozenge$}} 
    & 16*A100 
    & full-parameter SFT of LLaVA-Next-7B on SuperRS-VQA and HighRS-VQA with 81,367 QA pairs; expands visual grids from 7$\times$7 to 24$\times$24 for 8K inputs; achieves about 24$\times$ token compression; trained with ZeRO-2 for 1 epoch \\

 \\

    \bottomrule  
\end{tabular}%
\label{tab:modelTraining}%
\end{table*}

\begin{table*}[!t]  
  \centering  
  \caption{Instruction tuning data sources and processing methods of RS-MLLMs.}  
  \tiny
  \renewcommand{\arraystretch}{1.7} 
  \begin{tabular}{|>{\scriptsize}m{2.7cm}|>{\scriptsize}m{15cm}|}  
    \hline  
    \textbf{Model} & \textbf{Dataset Sources/ Processing/ Split} \\
    \hline  
    \makecell{RSGPT \\ \citep{Hu2023RSGPTAR}} & 
    RSICap: 2,585 high-quality RS image-text pairs; 5 RS experts annotate 2,500 images from DOTA-v1.5   \\
    \hline  
    
    \makecell{GeoChat \\ \citep{GeoChat}} & 
    Short captions generated through attribute extraction and expression synthesis using textual templates; 10,000 images for complex question answers and 30,000 images for detailed scene descriptions; Source: DOTA, DIOR, FAIR1M, LRBEN, Floodnet, NWPU-RESISC-45, SAMRS;  Multi-round Q\&A pairs using Vicuna-v1.5 with system instructions and prompts: 306,000 pairs for training; 12,000 pairs for testing \\
    \hline  
    
    \makecell{SkyEye-968k \\ \citep{Zhan2024SkyEyeGPTUR}} &   
    \textbf{Single-Task Image-Text.} Five captioning datasets: RSICD, RSITMD, UCM-Captions, Sydney-Captions, NWPU-Captions; UAV Video captioning dataset: CapERA; Three VQA datasets: RSIVQA, RSVQA-LR, RSVQA-HR, ERA-VQA (generated based on ERA); Two VG datasets: RSVG and DIOR-RSVG; Created phrase grounding dataset (RSPG).
    \textbf{Multitask Conversation.} 
    UCM-Conversa, Sydney-Conversa: Combined captioning and VQA datasets for dialogues; DIOR-Conversa: Integrates visual grounding, phrase grounding, and referring expression generation using the DIOR-RSVG and DIOR datasets; DOTA-Conversa: supports VQA and phrase grounding tasks using RSIVQA and the DOTA object detection (OD) dataset.  
    \\
    \hline

    \makecell{EarthGPT, \\ MMRS-1M \\ \citep{EarthGPT}} &   
    \textbf{Coarse-grained Conversation.} Five tasks and three visual modalities; Eight classification datasets: AID, EuroSAT, NWPU-RESISC45, UCMerced-LandUse, WHU-RS19, RSSCN7, ship classification datasets FGSCR-42 and DSCR; Five IC datasets: Sydney-Captions, RSICD, NWPU-Captions, RSITMD, and UCM-Captions; Four VQA datasets: Floodnet, RSVQA-LR, RSIVQA, and CRSVQA.   
    \textbf{Fine-grained Conversation.} Eight optical OD datasets: DIOR, DOTA, FAIR1M, HRRSD, NWPUVHR10, RSOD, UCAS-AOD, and VisDrone; Three SAR OD datasets: AIR-SARShip-2.0, HRISD, SSDD; Six infrared OD datasets: HIT-UAV, Sea-shipping, Infrared-security, Aerial-mancar, Double-light-vehicle, and oceanic ship; visual grounding and region-level caption: DIOR-RSVG.  
    \\
    \hline

    \makecell{LHRS-Bot \\ \citep{LHRS-Bot}} &   
    \textbf{LHRS-Align-1150k.} 1.15 million image-text pairs from Google Earth and OSM; Vicuna-v1.5-13B for caption generation, including metadata like resolution and location.  
    \textbf{LHRS-Instruct-39.8k.} Tailored for RS image understanding and complex visual reasoning; comprises 2.9K and 25K high-quality pairs sourced from RSITMD and NWPU, respectively, generating conversations with Vicuna-v1.5-13B and GPT-4. Additionally, 15K representative images were selected from LHRS-Align, with bounding boxes calculated using OSM, including 0.9K visual reasoning examples, 4K detailed descriptions, and 7K conversation samples.  
    \textbf{Multitask Instruction Dataset.} Collecting public RS datasets and creating manual instruction templates.  
    \\
    \hline
    
    \makecell{RS-LLaVA \\ \citep{RS-LLaVA}} &   
     \textbf{RS-Instructions Dataset.} Mixing existing data with two captioning datasets: UCM-caption and UAV dataset; Two VQA datasets: RSVQA-LR and RSIVQA-DOTA; The same training and testing split as the original datasets; 7058 samples, with 5506 samples in the training set and 1552 samples in the test set  \\
    \hline

    \makecell{H\textsuperscript{2}RSVLM \\  \citep{pang2025vhm}} &   
    \textbf{HqDC-1.4M.}  
     1.4M detailed image captions; Sources: Million-AID, fMoW, CVUSA, CVACT, LoveDA; ``gemini-1.0-pro-vision`` API for descriptions; Manual verified for accuracy, with 10.9\% identified as incorrect.  
     \textbf{HqDC-Instruct, 30k RS images.}  
     26k for conversation, 4k for reasoning; Sources: DOTA-v2, FAIR1M training sets; Gemini-Vision for captions; Gemini language model for multi-turn conversation and reasoning data.  
    \textbf{RSSA, 44k.}  
    Tasks: Presence, Color, AbsPos, RelPos; Yes-or-No, Open-ended, Multiple Choice, Unswervable; Sources: DOTA-v2, Fair1M.  
    \textbf{RS-Specialized-Instruct, 29.8k.}  
    Tasks: GSDEst, ImgType, ObjMeas, MlLc, BFV; Sources: BANDON, MtS-WH, DOTA-v2, MSAR, FAIR1M, fMoW, CrowdAI, GID, FBP, DeepGlobe.  
    \textbf{RS-ClsQaGrd-Instruct Multitask dataset, 78k.}  
    Tasks: SC, VQA, VG; Sources: RSVQA-LR, fMoW, METER-ML, NWPU, RSITMD, UCM, DIOR-RSVG.  
    \\
    \hline

   \makecell{SkySenseGPT \\ \citep{SkySenseGPT}} &   
     \textbf{FIT-RS, 1800k+ samples.}  
    Created from the graph dataset STAR, simple tasks involve compiling instructions from complete scene graph data using TinyLLaVA-3.1B and GPT-4/GPT-3.5; complex tasks include modifying information within the graph and mixing different tasks to generate multi-turn conversation data. The dataset includes negative samples, with instructions to refuse to answer when presented with background images. The data is divided into training, validation, and testing sets at a ratio of 6:2:2.  
    \textbf{Additional datasets for training, 365k samples.} Three SC datasets: NWPU, UCM, and RSITMD; Three VQA datasets: EarthVQA, Floodnet-VQA, and RSVQA-LR; Three OD datasets: DOTA-v2.0, DIOR, and FAIR1M.  
    \\
    \hline

    \makecell{RS-GPT4V \\ \citep{rsgpt}} &   
    \textbf{Five captioning datasets}: NWPU-Captions, RSICD, RSITMD, Sydney-Captions, UCM-Captions; \textbf{Four VQA datasets}: RSVQA-LR, RSVQA-HR, FloodNet, RSIVQA; One VG dataset: DIOR-RSVG; \textbf{One region-level captioning dataset}: DIOR-RSVG; \textbf{Multi-turn Conversation}: RS-GPT4V-Instruct;\textbf{ Detailed Description}: RS-GPT4V-Instruct; Using GPT-4V with specific instructions; 91,937 training images with 991,206 question-answer pairs and 15,999 test images with 258,419 question-answer pairs  \\
    \hline

    \makecell{Falcon \\ \citep{yao2025falcon}} &  \textbf{Falcon SFT.} A large-scale multi-task RS instruction-tuning dataset with approximately 78M samples from 5.6M multi-resolution and multi-view RS images. It integrates 67 open-source RGB RS datasets and supports 14 image-, region-, and pixel-level tasks, including classification, counting, VQA, captioning, detection, visual grounding, segmentation, and change detection.   \\
    \hline

    \makecell{Geo-R1 \\ \citep{zhang2026geo}} & \textbf{Few-shot REU datasets.} Geo-R1 focuses on few-shot geospatial referring expression understanding, including REC, OVD, and GRES. It constructs VRSBench-FS, NWPU-FS, and EarthReason-FS from existing RS benchmarks, and applies reinforcement fine-tuning based on GRPO/DAPO with task-specific rewards, such as BBoxIoU, mAP, and MaskGIoU.   \\
    \hline

    \makecell{ZoomEarth \\ \citep{liu2026zoomearth}} &  \textbf{LRS-GRO.} An ultra-high-resolution RS VQA benchmark for active perception, containing 1,224 images, 3,592 bounding boxes, and 13,245 questions. It covers 17 global-, region-, and object-level question types. ZoomEarth is trained with SFT and GRPO, using CoT-style cropping annotations and a Region-Guided reward for adaptive cropping--zooming.   \\
    \hline

    \makecell{GeoLLaVA-8K \\ \citep{wang2026geollava}} & \textbf{SuperRS-VQA and HighRS-VQA.} Two ultra-high-resolution RS VQA datasets containing 81,367 VQA pairs in total and covering 22 real-world dialogue tasks. GeoLLaVA-8K is trained by full-parameter SFT based on LLaVA-Next-7B and introduces Background Token Pruning and Anchored Token Selection to process up to 8K-resolution inputs.    \\
    \hline  
\end{tabular}%
  \label{tab:datasets}%
\end{table*}

\subsection{RSISU Requirements}

Unique requirements of RSISU. Successfully adapting MLLMs for RS applications necessitates addressing the following three primary aspects. 1) \textbf{Data characteristics}: MLLMs must be equipped to process multi-resolution, multimodal, multi-platform, and multi-temporal imagery from various viewpoints. This capability involves handling diverse atmospheric conditions and integrating temporal data for change detection. Additionally, these models must incorporate geospatial information to enhance contextual understanding. 2) \textbf{Task-specific requirements}: RSISU applications require fine-grained object recognition, particularly for small and diverse objects and advanced spatial reasoning to interpret complex geographical layouts and relationships. RS-MLLMs must be optimized for specialized tasks, such as land use classification and crop monitoring while ensuring explainable outputs to support decision-making processes. 3) \textbf{Domain knowledge integration}: RS-MLLMs need to possess a comprehensive understanding of RS-specific features, terrain types, and man-made structures. Furthermore,  these models must comprehend the broader geographical and environmental context of observed features.

\subsection{Remote Sensing FMs}

As summarized in Table~\ref{tab:rsEncoder}, the development of RS FMs exhibits a clear progression from single-modality visual pretraining to multi-modal and vision-language representation learning. Early RS FMs mainly focus on visible optical imagery, where models such as Geograph~\citep{Ayush2020GeographyAwareSL}, RSP~\citep{Wang2022AnES}, SatMAE~\citep{Cong2022SatMAEPT}, ScaleMAE~\citep{Reed2022ScaleMAEAS}, RingMo~\citep{Sun2023RingMoAR}, Billion~\citep{Cha2023ABF}, and GFM~\citep{Mendieta2023GFMBG} learn transferable visual representations from large-scale RS images. Subsequently, RS FMs have been extended to more diverse Earth observation modalities. SpectralGPT~\citep{spectralgpt}, HyperSIGMA~\citep{hypersigma}, and HyperFree~\citep{hyperfree} target spectral imagery, while SARJEPA~\citep{SARJEPA}, SARATR-X~\citep{SARATRX}, and SARMAE~\citep{sarmae} focus on SAR representation learning. More recent models, such as SkySense~\citep{skysense}, TerraFM~\citep{terrafm}, and SkySense V2~\citep{skysensev2}, further pursue unified representations across multiple RS modalities, reflecting the heterogeneous and multi-source nature of Earth observation.

Another important trend is the transition from purely visual representation learning to vision-language pretraining. Methods such as SkyCLIP~\citep{skyscript}, RemoteCLIP~\citep{Liu2023RemoteCLIPAV}, GRAFT~\citep{GRAFT}, GeoRSCLIP~\citep{Zhang2023RS5MAG}, SARVLM~\citep{sarvlm}, SARCLIP~\citep{sarclip-isprs}, and HyperCap~\citep{HyperCap} align RS imagery with natural-language descriptions, enabling cross-modal retrieval, zero-shot recognition, and downstream RS vision-language tasks. These visual and vision-language FMs provide important encoder candidates for RS-MLLMs. However, most existing RS FMs are still designed for representation learning or cross-modal alignment, rather than interactive reasoning, multi-image understanding, instruction following, and complex RSISU tasks. Therefore, how to effectively integrate RS-specific foundation encoders with powerful LLM backbones remains a key issue for building more capable RS-MLLMs.

\subsection{Remote Sensing MLLMs}
RS-MLLMs~\citep{qwen3vl,glm46v,li2026graph} continue to evolve rapidly, as illustrated in Figure~\ref{fig:RS-MLLMs}. RSGPT represents a pioneering integration of LLMs with RS tasks, focusing on IC and VQA \citep{Hu2023RSGPTAR}. By adopting the InstructBLIP model, RSGPT demonstrates notable performance in scene description tasks and introduces a dense captioning dataset known as RSICap. However, despite its strengths, RSGPT requires task-specific tuning, which limits its generalizability. It also faces challenges in handling diverse tasks, including region-level reasoning, and specific tasks such as classification, detection, and visual grounding.

GeoChat extends conversational capabilities within the remote sensing domain by enabling region-specific dialogues and visual grounding \citep{GeoChat}. Built upon the LLaVA1.5 architecture, GeoChat effectively handles both image-level and region-specific queries, grounding objects in images using spatial coordinates. However, despite advancements in spatial reasoning, GeoChat encounters challenges in tasks requiring fine-grained recognition, such as detailed object recognition and counting.

SkyEyeGPT unifies visual-language tasks at both image and region levels, utilizing an architecture based on MiniGPT-v2 \citep{Zhan2024SkyEyeGPTUR}. It employs a two-stage tuning method to enhance instruction-following and multi-turn dialogue capabilities across varying levels of granularity. Notably,  SkyEyeGPT is the first model to support RS video captioning tasks, marking a significant step in expanding RS-MLLM functionalities.

EarthGPT focuses on the unification of multi-sensor RS tasks and the enhancement of RS expert knowledge \citep{EarthGPT}. By leveraging the large-scale dataset MMRS-1M, which comprises diverse sensor images such as optical, synthetic aperture radar (SAR), and infrared, EarthGPT emphasizes cross-modal comprehension. This inclusion of SAR and infrared images broadens the model's input capabilities and supports more versatile RS applications.

LHRS-Bot employs an innovative multi-level vision-language alignment strategy complemented by a curriculum learning approach. It introduces the LHRS-Align and LHRS-Instruct datasets, leveraging geographical information to advance understanding and reasoning in RSISU tasks. This methodology enhances the model's ability to process complex spatial information effectively.

RS-CapRet specializes in IC and text–image retrieval for RS tasks \citep{silva2024largelanguagemodelscaptioning}. It utilizes a straightforward training procedure that aligns a visual encoder with a language model and can manage interleaved image-text dialogues. While innovative, RS-CapRet has a narrower focus compared to other models that address a broader range of tasks. RS-LLaVA emphasizes multitasking capabilities within RS imagery, employing a low-rank adaptation approach primarily focused on IC and VQA \citep{RS-LLaVA}.

H\textsuperscript{2}RSVLM stands out by introducing trustworthiness as a central concept in RS models, enhancing the honesty and helpfulness of RS-MLLMs through targeted datasets. It tackles the prevalent hallucination problem by developing resources that strengthen the model’s self-awareness and its ability to decline responses to unanswerable questions.

SkySenseGPT advances relation reasoning and comprehension tasks through a large-scale instruction tuning dataset \citep{SkySenseGPT}. Utilizing graphs to improve perception and relational reasoning in RS imagery addresses the template limitations of earlier models. Its significant contribution lies in managing complex relationships critical for fine-grained scene understanding, moving from basic relation reasoning to image-level and region-level graph generation. RS-GPT4V introduces the RS-GPT4V dataset to unify multimodal tasks and hierarchical instruction descriptions, focusing on reasoning and detailed scene comprehension \citep{rsgpt}. By employing strategies such as local and global hierarchical instruction descriptions and multi-turn question-answer pairs, RS-GPT4V captures fine-grained information and implicit knowledge across diverse RS tasks, surpassing existing datasets in coverage and capability.

Falcon presents a holistic RS vision-language foundation model with unified prompt-based outputs across image-, region-, and pixel-level tasks \citep{yao2025falcon}. Supported by the large-scale Falcon SFT dataset, which contains about 78 million samples from 5.6 million RS images, Falcon is capable of handling 14 tasks, including IC, VQA, counting, captioning, detection, grounding, segmentation, and change detection. Compared with previous RS-MLLMs that mainly focus on a limited set of image- or region-level tasks, Falcon significantly expands the functional coverage of RS vision-language modeling. However, despite its comprehensive task design, Falcon is still mainly trained and evaluated within predefined task templates, and its open-ended reasoning ability in complex real-world RS scenarios requires further investigation.

\begin{table*}[!t] 
  \centering  
  \renewcommand{\arraystretch}{1.2}
  \caption{Evaluation methods and datasets of SOTA RS-MLLMs.}  
  \scriptsize
  \begin{tabular}{  
    >{\centering\arraybackslash}m{2.5 cm}  
    >{\raggedright\arraybackslash}m{15cm} 
  }  
    \toprule  
    \textcolor[rgb]{ .024,  .024,  .027}{\textbf{Model}} & \textcolor[rgb]{ .024,  .024,  .027}{\textbf{Evaluation Methods}} \\
    \midrule
    
    \makecell{RSGPT \\ \citep{Hu2023RSGPTAR}} &   
    \textbf{Non-ZS Evaluation:}\newline   
    (1) RSIEval: manual annotation of 100 images from the DOTA-v1.5 validation set, 100 image-caption pairs with one caption per image, and 936 diverse image-question-answer triplets with an average of 9 questions per image (four categories: object-related, image-related, scene-related, reasoning-related); (2) testing fine-tuned RSGPT: For IC, the UCM-captions, Sydney-captions, and RSIC datasets are used; for VQA, the RSVQA-LR and RSVQA-HR datasets are employed. \\
    \midrule
    
    \makecell{GeoChat \\ \citep{GeoChat}} &   
    \textbf{ZS Evaluation:}\newline   
    (1) SC: AID, UCMerced (2) VQA: RSVQA-HRBEN (47k); RSVQA-LRBEN (7k); Area and count excluded\newline  
    \textbf{Non-ZS Evaluation:}\newline   
    VG and captioning: newly created dataset using SAMRS validation set (765 ref, 758 grounding, 555 descriptions), i.e., the test set of GeoChat-318k \\
    \midrule
    
    \makecell{SkyEyeGPT \\ \citep{SkySenseGPT}} &   
    \textbf{Non-ZS Evaluation:}\newline   
    (1) Captioning: UCM-captions (2) VQA: RSVQA-LR test set and RSVQA-HR test set (2) (3) VG: RSVG and DIOR-RSVG (4) video captioning: CapERA \\
    \midrule 
    
    \makecell{EarthGPT \\ \citep{EarthGPT}}  &   
    \textbf{ZS Evaluation:}\newline   
    (1) SC: all images from CLRS and NaSC-TG (2) VQA: RSVQA-HR, test set (2) (3) OD: MAR\newline  
    \textbf{Non-ZS Evaluation:}\newline   
    (1) SC: NWPU-RESISC45 (2) IC: test set of the NWPU-Caption (3) VG: test set of DIOR-RSVG \\
    \midrule  
    
    \makecell{LHRS-Bot \\ \citep{LHRS-Bot}}  &   
    \textbf{ZS Evaluation:}\newline   
    (1) LHRS-Bench  Benchmark: Comprises 690 single-choice questions across 11 dimensions. The images are sourced from Google Earth, and no public RS datasets have been utilized. This benchmark is manually crafted to ensure precision and reliability and includes a total of 108 images paired with 690 questions.\newline  
    \textbf{Non-ZS Evaluation:}\newline   
    (1) SC: METER-ML, fMoW (2) VQA: RSVQA-LR and RSVQA-HR (3) VG: RSVG and DIOR-RSVG \\
    \midrule  
    
    \makecell{RS-LLaVA \\ \citep{RS-LLaVA}}  &   
    \textbf{Non-ZS Evaluation:}\newline   
    (1) IC: UCM-caption and UAV dataset (2) VQA: RSVQA-LR, and RSIVQA-DOTA \\
    \midrule  
    
    \makecell{H\textsuperscript{2}RSVLM \\ \citep{pang2025vhm}}   &   
    \textbf{ZS Evaluation:}\newline   
    (1) SC: SIRI-WHU, AID, and WHU-RS19 (2) VQA: RSVQA-HR\newline  
    \textbf{Non-ZS Evaluation:}\newline   
    (1) Test set of RSSA (2) RS-ClsQaGrd-Instruct datasets \\
    \midrule
    
    \makecell{SkySenseGPT \\ \citep{SkySenseGPT}}  &   
    \textbf{ZS Evaluation:}\newline   
    (1) SC: SIRI-WHU, AID, WHU-RS19, AID-multi (2) VQA: RSVQA-HR; Area and count excluded\newline  
    \textbf{Non-ZS Evaluation:}\newline   
    (1) IC, RC, VQA, SC, complex comprehension, multi-turn conversation: test set of each task in FIT-RS (2) VQA: RSVQA-LR; Area and count excluded (3) FIT-RSRC based on STAR single-choice questions with 4 types (relationship, object, subject, existence) \\
    \midrule  
    
    \makecell{RS-GPT4V \\ \citep{rsgpt}} &   
    \textbf{Non-ZS Evaluation:}\newline   
    (1) IC: UCM-Captions, RSICD, Sydney-Captions, NWPU-captions (2) VQA: RSVQA-LR and RSVQA-HR (3) VG: DIOR-RSVG (4) RS-GPT4V-Instruct dataset \\
    \midrule

    \makecell{Falcon \\ \citep{yao2025falcon}} &   
    \textbf{Non-ZS Evaluation:}\newline   
    (1) 14 RS vision-language tasks, including IC, VQA, counting, captioning, detection, grounding, segmentation, and change detection\newline
    \textbf{ZS Evaluation:}\newline
    (1) UCM-Captions (2) MAR20 (3) NWPU-VHR-10 (4) GID15 (5) CCD (6) WHU-CD \\
    \midrule

    \makecell{Geo-R1 \\ \citep{zhang2026geo}} &   
    \textbf{Few-shot Evaluation:}\newline   
    (1) REC: VRSBench-FS / VRSBench test set (2) OVD: NWPU-FS / NWPU VHR-10 test set (3) GRES: EarthReason-FS / EarthReason validation and test sets\newline
    \textbf{Cross-dataset ZS Evaluation:}\newline
    (1) REC: DIOR-RSVG (2) GRES: RRSIS-D \\
    \midrule

    \makecell{ZoomEarth \\  \citep{liu2026zoomearth}} &   
    \textbf{Non-ZS Evaluation:}\newline   
    (1) LRS-GRO test set: UHR RS VQA benchmark with global-, region-, and object-level questions\newline
    \textbf{ZS Evaluation:}\newline
    (1) MME-RealWorld-RS (2) XLRS-Bench (3) GeoLLaVA-8K \\
    \midrule

    \makecell{GeoLLaVA-8K \\ \citep{wang2026geollava}}  &
    \textbf{ZS Evaluation:}\newline   
    (1) XLRS-Bench: UHR RS VQA benchmark covering perception and reasoning tasks\newline
    \textbf{Generalization Evaluation:}\newline
    (1) LRS-VQA converted to multiple-choice format \\
    \bottomrule  
  \end{tabular}%
  \label{tab:evaluation}%
\end{table*}

Geo-R1 further explores the potential of reasoning- oriented reinforcement fine-tuning in RS-MLLMs, especially under few-shot geospatial referring expression understanding scenarios \citep{zhang2026geo}. Different from conventional SFT-based paradigms, Geo-R1 adopts GRPO-based reinforcement fine-tuning to encourage explicit reasoning chains before producing spatial predictions. By constructing VRS- Bench-FS, NWPU-FS, and EarthReason-FS, Geo-R1 provides a standardized few-shot evaluation protocol for REC, OVD, and GRES tasks. Its main contribution lies in improving spatial grounding and cross-dataset generalization with limited annotated samples. However, its application scope is mainly concentrated on referring expression understanding, while broader multimodal RS tasks such as captioning, general VQA, and multi-sensor interpretation remain less explored.

ZoomEarth addresses the challenge of ultra-high resolution RS image understanding by introducing an active perception paradigm \citep{liu2026zoomearth}. Instead of passively processing the whole image at a fixed resolution, ZoomEarth adaptively invokes a cropping--zooming tool to revisit informative regions and conduct fine-grained reasoning. Built upon the LRS-GRO benchmark, it covers global-, region-, and object-level questions and employs SFT followed by GRPO with a Region-Guided reward. This design enhances the model's ability to localize regions of interest and reason over detailed visual contents in UHR imagery. Nevertheless, ZoomEarth relies on tool-based region exploration, and its effectiveness may depend on the accuracy of region selection and the quality of cropped visual evidence.

GeoLLaVA-8K focuses on scaling RS-MLLMs to ultra-high-resolution inputs up to 8K$\times$8K \citep{wang2026geollava}. To address the lack of UHR vision-language data, it introduces SuperRS-VQA and HighRS-VQA, two high-resolution RS VQA datasets covering diverse real-world dialogue tasks. Meanwhile, to mitigate the token explosion problem caused by large image sizes, GeoLLaVA-8K proposes Background Token Pruning and Anchored Token Selection, preserving semantically important tokens while reducing computational cost. This enables effective UHR RS understanding and achieves strong performance on XLRS-Bench. However, its current focus is mainly on optical satellite imagery and VQA-style tasks, leaving broader sensor modalities and more diverse RS reasoning tasks for future exploration.

\subsection{Remote Sensing Agents}
\label{subsection:rs_agent}
With the rapid development of agent technologies~\citep{liu2024llava,yang2025magma,li2026same} in the general computer vision community, remote sensing agents have also begun to attract increasing research attention. Remote sensing agents represent a recent extension of RS-MLLMs from passive vision-language interaction toward executable geospatial intelligence. Unlike conventional RS-MLLMs that mainly answer questions or generate descriptions from input images, RS agents aim to understand user instructions, plan multi-step workflows, invoke external tools, and integrate intermediate results for final decision making. This paradigm is particularly important for RSISU because many real-world remote sensing tasks require not only visual perception, but also geospatial data access, object detection, segmentation, change detection, spatial computation, retrieval, and domain-specific analysis.

\textbf{Tool-augmented RS agents.}
The first line of development focuses on tool-augmented RS agents. Instead of relying only on the internal perception and reasoning ability of a single MLLM, these systems use an LLM or MLLM as the central controller to decompose user instructions and call specialized tools. RS-Agent introduces a central controller, dynamic toolkit, solution space, and knowledge space to automate remote sensing tasks, while task-aware retrieval and DualRAG improve tool selection and domain knowledge grounding \citep{xu2024rsagent}. GeoLLM-Engine further builds a realistic environment for geospatial copilots by incorporating geospatial APIs, dynamic maps, user-interface states, and external multimodal knowledge sources \citep{singh2024geollm}. ThinkGeo evaluates tool-augmented agents on structured remote sensing tasks involving optical and SAR imagery, emphasizing multi-step planning and tool execution \citep{shabbir2025thinkgeo}. These studies show that RS agents can connect MLLMs with detectors, segmentation models, change detection algorithms, GIS functions, and knowledge bases, thereby extending RS-MLLMs from image-level understanding to remote sensing workflows.

\textbf{Benchmark and process-level evaluation.}
A second line focuses on benchmark construction and process-level evaluation for RS agents. Existing RS-MLLM benchmarks usually evaluate final answers on static image-text pairs, which are insufficient for agentic systems that interact with tools and dynamic platforms. GeoLLM-QA argues that RS agent evaluation should consider long sequences of verbal, visual, and click-based actions on real remote sensing platforms, where user intent may depend on map locations, selected regions, and UI states \citep{singh2024evaluating}. ThinkGeo similarly evaluates not only final-answer correctness, but also step-wise execution quality, tool accuracy, and planning consistency, which are critical for diagnosing failures in long-horizon RS workflows \citep{shabbir2025thinkgeo}. Earth-Agent and Earth-Bench further strengthen this direction by evaluating both reasoning trajectories and final results for EO tasks requiring domain-specific tools and multi-step quantitative analysis \citep{feng2026earthagent}. These works suggest that future RS agent benchmarks should measure whether the agent selects the right tools, follows reasonable intermediate steps, and produces verifiable conclusions.

\textbf{Multi-agent geospatial copilots.}
A third direction is multi-agent geospatial copilots. Complex RS workflows often involve heterogeneous data sources, multiple analytical stages, and different types of expertise, making it difficult for a single monolithic agent to handle all subtasks robustly. GeoLLM-Squad addresses this issue by separating agentic orchestration from task-specific geospatial problem solving and delegating subtasks to specialized agents \citep{lee2025multiagent}. Such a design naturally supports roles such as planner, retriever, vision worker, GIS worker, and verifier, and can improve scalability for applications such as urban monitoring, forestry protection, climate analysis, and agricultural studies. This multi-agent paradigm indicates that RS agents may evolve from single-controller systems toward collaborative geospatial copilot teams.

\textbf{Domain-specific and professional RS agents.}
A fourth direction moves RS agents toward domain-specific and professional scenarios. MineAgent targets remote-sensing mineral exploration by combining geological knowledge, hyperspectral information, and multi-image reasoning \citep{yu2024mineagent}. REMSA focuses on automatic remote sensing foundation model selection, using an LLM agent to interpret user requirements and rank suitable RSFMs from a structured model database \citep{chen2025remsa}. ShapefileGPT extends agentic geospatial analysis to vector GIS data by using a multi-agent framework for automated Shapefile processing and spatial queries \citep{lin2025shapefilegpt}. These studies suggest that future RS agents will increasingly move from general image understanding toward professional workflows such as mineral exploration, disaster assessment, foundation model recommendation, and vector GIS analysis.

\textbf{Active tool exploration and unified multimodal reasoning.}
More recent studies further address scalability in tool organization and multimodal reasoning. RS-Claw tackles the context burden of large tool libraries by organizing tools into hierarchical skill trees and enabling progressive active tool exploration, thereby reducing token overhead while preserving access to critical tools \citep{liu2026rsclaw}. In parallel, RingMo-Agent follows a model-centric route by building a unified multi-platform and multi-modal RS foundation model over optical, SAR, and infrared imagery from satellite and UAV platforms \citep{hu2025ringmoagent}. Overall, the development path of RS agents is becoming clear: from single-model RS-MLLMs, to tool-augmented agents, to realistic platform evaluation, multi-agent collaboration, domain-specific applications, and active tool exploration. Despite this progress, RS agents still face challenges in reliable tool selection, long-horizon reasoning, trajectory verification

\begin{table*}[!t]
    \centering
    \caption{Contamination-risk audit of the benchmarks considered in this study.}
    \label{tab:contamination_audit}
    \scriptsize
    \renewcommand{\arraystretch}{1.5}
    \setlength{\tabcolsep}{3.5pt}
    \begin{tabular}{
    p{2.5cm}
    p{2.0cm}
    p{6.0cm}
    p{4.5cm}
    p{2cm}}
    \hline
    \textbf{Benchmark} &
    \textbf{Public date} &
    \textbf{Identified overlap} &
    \textbf{Evaluation interpretation} &
    \textbf{Risk} \\
    \hline
    
        \makecell[tl]{AID\\ \citep{AID}\\WHU-RS19\\ \citep{WHU-RS19}\\EuroSAT\\ \citep{Eurosat}} &
        \makecell[tl]{2017-04\\ \\ 2010-08 \\ \\ 2019-06} &
        Some RS-MLLMs explicitly use the corresponding classification datasets
        for instruction tuning, while the training corpora of other models remain
        partially undisclosed. &
        Benchmark-aware inference; not strict ZS for explicitly trained models. &
        Medium / High \\
        
        \makecell[tl]{AID-multi\\ \citep{AID_multi}} &
        2020-02 &
        AID-multi is derived from AID; models trained on AID may therefore have
        image-level overlap without using its multi-label annotations. &
        Benchmark-aware evaluation. &
        Medium / High \\
        
        \makecell[tl]{RSVQA-HR\\ \citep{RSVQA}} &
        2020-03 &
        RSVQA-HR is explicitly included in the instruction-tuning data of several
        RS-MLLMs, including SkyEyeGPT and RS-GPT4V. &
        In-domain for overlapping models; otherwise no-adaptation inference. &
        Medium / High \\
        
        \makecell[tl]{RSVG\\\citep{SunRSVG}\\DIOR-RSVG\\\citep{DIOR-RSVG}\\VRSBench\\\citep{VRSBench}}  &
        \makecell[tl]{2022-10 \\ \\ 2022-10 \\ \\2024-06 }&
        Several RS-MLLMs use the corresponding training splits, few-shot subsets,
        or closely related grounding data. &
        In-domain, few-shot, or cross-dataset evaluation, depending on the model. &
        Medium / High \\
        
        \makecell[tl]{LHRS-Bench\\\citep{LHRS-Bot} }&
        2024-02 &
        The benchmark uses separately collected Google Earth images, and no
        explicit training overlap has been identified. &
        Strict ZS for eligible pre-release models; benchmark-aware for later models. &
        Low / Medium \\
        
        \makecell[tl]{FIT-RSRC\\\citep{SkySenseGPT} } &
        2024-06 &
        SkySenseGPT is evaluated under transfer learning, while several recent
        models were released after the benchmark became public. &
        Transfer learning for SkySenseGPT; benchmark-aware inference for others. &
        Medium / High \\
        
        \makecell[tl]{XLRS-Bench\\ \citep{xlrs_bench}} &
        2025-03 &
        The strongest evaluated models were released after the benchmark. Its
        images partly originate from public RS datasets, but direct test-annotation
        exposure has not been identified. &
        Benchmark-aware evaluation rather than contamination-free ZS. &
        Medium \\
    
    \hline
    \end{tabular}
    
    \vspace{1mm}
    \parbox{0.98\textwidth}{
    \footnotesize
    \textit{Risk definition:} Low indicates no identified overlap and an eligible
    pre-release checkpoint; Medium indicates post-release evaluation or incomplete
    training-data disclosure without confirmed overlap; High indicates explicit
    use of the benchmark, its source images, training split, few-shot subset, or
    task annotations.
    }
\end{table*}

\subsection{Instruction Tuning Datasets}
In this paragraph, as shown in Table~\ref{tab:datasets}, the instruction tuning datasets used in representative RS-MLLMs are illustrated. Existing datasets are generally constructed by reorganizing public RS benchmarks and enriching them with captions, question-answer pairs, grounding annotations, or multi-turn conversations. These datasets cover diverse tasks, including scene classification, image captioning, VQA, visual grounding, object detection, region-level captioning, reasoning, segmentation, and change detection. Recent works further extend instruction tuning toward fine-grained relation reasoning, ultra-high-resolution perception, and active visual interaction.

Among these datasets, LHRS-Bot\citep{LHRS-Bot} introduces LHRS-Align and LHRS-Instruct. LHRS-Align contains 1.15M image-text pairs collected from Google Earth and OSM, while LHRS-Instruct provides RS-oriented instruction data for image understanding and complex visual reasoning. SkySenseGPT~\citep{SkySenseGPT} constructs FIT-RS with more than 1.8M samples based on scene graph data, covering both simple and complex comprehension tasks. It further incorporates additional training data from scene classification, VQA, and object detection datasets. RS-GPT4V~\citep{rsgpt} integrates captioning, VQA, visual grounding, region-level captioning, multi-turn conversation, and description data, using GPT-4V to generate instruction-following samples.

More recent datasets show a clear trend toward scale, task diversity, and high-resolution understanding. Falcon~\citep{yao2025falcon} builds a large-scale SFT dataset with approximately 78M samples from 5.6M RS images and supports 14 tasks across image-, region-, and pixel-level understanding. Geo-R1~\citep{zhang2026geo} focuses on few-shot geospatial referring expression understanding and constructs task-specific datasets for REC, OVD, and GRES. ZoomEarth~\citep{liu2026zoomearth} introduces LRS-GRO for ultra-high-resolution RS VQA with active cropping--zooming annotations, while GeoLLaVA-8K~\citep{wang2026geollava} builds SuperRS-VQA and HighRS-VQA to support 8K-resolution RS image understanding. These developments indicate that RS instruction tuning datasets are evolving from task-specific collections toward large-scale, fine-grained, and high-resolution multimodal corpora.

\begin{table*}[t]  
  \centering 
  \scriptsize
  \caption{General representative CV-MLLMs in our study.}
  \renewcommand{\arraystretch}{1.2}
  \setlength{\tabcolsep}{0.1pt}
  \begin{tabular*}{\hsize}{@{\extracolsep{\fill}}llll}  
    \hline  
    \textbf{MLLMs} & \textbf{LLMs} & \textbf{Source} & Website     \\
    \hline  
    Qwen-VL-Chat &  Qwen-7B          & \citep{bai2023qwenvlversatilevisionlanguagemodel} & \url{https://huggingface.co/Qwen/Qwen-VL-Chat}   \\
    Kosmos-2     & TCLM-1.3B         & \citep{peng2023kosmos2groundingmultimodallarge} & \url{https://huggingface.co/microsoft/kosmos-2-patch14-224} \\
    Next\_chat   & Vicuna-1.5-7B     & \citep{zhang2023nextchatlmmchatdetection}  &  \url{https://huggingface.co/AoZhang/nextchat-7b-336}  \\
    Qwen2-VL     &  Qwen-7B          & \citep{Qwen2VL} & \url{https://huggingface.co/collections/Qwen/qwen2-vl}  \\
    Cogvlm-VG    &  Vicuna v1.5-7B   & \citep{wang2024cogvlmvisualexpertpretrained} & \url{https://huggingface.co/THUDM/cogvlm-grounding-generalist-hf}  \\
    CogVLM-Chat  &  Vicuna v1.5-7B   & \citep{wang2024cogvlmvisualexpertpretrained} & \url{https://huggingface.co/THUDM/cogvlm-grounding-generalist-hf} \\
    Llava-1.5    & Vicuna v1.5-7B    & \citep{liu2024improvedbaselinesvisualinstruction}  & \url{https://huggingface.co/llava-hf/llava-1.5-7b-hf}  \\
    Llava-1.6   & Mistral-7B        & \citep{llava16} & \url{https://huggingface.co/llava-hf/llava-v1.6-mistral-7b-hf}  \\
    TinyLLaVA   & Phi-2-2.7B        & \citep{zhou2024tinyllavaframeworksmallscalelarge} & \url{https://huggingface.co/tinyllava/TinyLLaVA-Phi-2-SigLIP-3.1B} \\
    Qwen3-VL   & Qwen3-7B             & \citep{qwen3vl} & \url{https://huggingface.co/collections/Qwen/qwen3-vl} \\
    InternVL3.5 & Qwen3-7B            & \citep{internvl3_5} & \url{https://huggingface.co/collections/OpenGVLab/internvl35} \\
    GLM-4.6V-flash   & GLM4-9B    & \citep{glm46v}   &\url{https://huggingface.co/zai-org/GLM-4.6V-Flash} \\
    \hline  
  \end{tabular*}  
  \label{tab:cvModels}  
\end{table*}

\subsection{Model Architecture and Fine-tuning}
\label{sec:architecture-fine-tuning}

As illustrated in Figure~\ref{fig:modelarchitecture}, a typical MLLM consists of a modality encoder, an input projector, an LLM backbone, and, in some cases, an output projector or modality generator for multimodal generation. The modality encoder extracts visual or multimodal features, the input projector maps them into the language embedding space, and the LLM backbone performs instruction following, reasoning, and response generation. Early RS-MLLMs were mainly built upon LLaMA and  Vicuna-based architectures due to their accessibility and compatibility with LLaVA-style visual instruction tuning \citep{liu2023llava}. More recently, stronger general-purpose MLLM backbones, such as Qwen-VL, Qwen2.5VL, Qwen3-VL, InternVL, and GLM-based models, have been increasingly adopted because of their improved multilingual understanding, visual reasoning, and high-resolution perception. This trend is also reflected in recent RS-MLLMs, where Geo-R1 and ZoomEarth are initialized from Qwen2.5VL, while GeoLLaVA-8K adopts LLaVA-Next for ultra-high-resolution image understanding.

The central challenge in MLLMs is to align heterogeneous modality features with the semantic space of LLMs while preserving their instruction-following and reasoning capabilities. Existing training pipelines usually include multimodal pretraining and instruction tuning. The former aligns visual representations with textual embeddings through the input projector, whereas the latter improves task generalization, interaction ability, and domain-specific reasoning. Most early RS-MLLMs rely on supervised fine-tuning (SFT) or parameter-efficient fine-tuning methods, such as LoRA, to reduce training cost. Recent methods further introduce reinforcement-based optimization to enhance reasoning and grounding. For example, Geo-R1 adopts GRPO-based reinforcement fine-tuning for geospatial referring expression understanding, while ZoomEarth combines SFT and GRPO with cropping--zooming tools for ultra-high-resolution image interpretation.

Table~\ref{tab:modelTraining} summarizes the architecture and tuning strategies of representative RS-MLLMs. Overall, early models commonly freeze the visual encoder and adapt the system through the input projector and LoRA-based LLM tuning. RSGPT is a lightweight example that only tunes the input projector, whereas H\textsuperscript{2}RSVLM fine-tunes the visual encoder, input projector, and LLM during pretraining. LHRS-Bot adopts a more elaborate multi-stage pipeline by combining alignment data, multitask instruction data, detailed descriptions, and selected public reasoning data. SkySenseGPT also incorporates additional public datasets for fine-grained RS instruction tuning. More recent RS-MLLMs show a clear shift from simple projector-level adaptation toward stronger foundation backbones and task-oriented training paradigms. Falcon performs large-scale SFT across multiple RS tasks, Geo-R1 introduces reinforcement fine-tuning for geospatial reasoning, ZoomEarth integrates active perception with cropping--zooming tools, and GeoLLaVA-8K improves high-resolution image understanding through token pruning and visual token selection. These developments indicate that RS-MLLMs are moving toward stronger backbones, high-resolution perception, reasoning-oriented training, and tool-augmented interaction.

\begin{table}[t]
  \centering
  \scriptsize
  \setlength{\tabcolsep}{0.2pt}
  \renewcommand{\arraystretch}{1.2}
  \caption{Comparative scene classification performance of RS-MLLMs and CV-MLLMs. Bold numbers denote the best performance among this study.}
    \begin{tabular*}{\hsize}{@{\extracolsep{\fill}}lccccc}
    \toprule
    \textbf{Model} & \textbf{AID} & \textbf{WHU19} & \textbf{EuroSAT} & \textbf{AID-multi} & \textbf{Source} \\
    \midrule
    LLaVA-1.5          & 31.1  & 54.6  & 26.1  &  /     & \multirow{7}{*}{\makecell{LHRSBot \\ \citep{LHRS-Bot}}} \\
    MiniGPTv2          & 33.0  & 64.8  & 38.6  &  /     &  \\
    Qwen-VL-Chat       & 55.3  & 72.3  & 26.4  &  /     &  \\
    InstructBLIP       & 29.5  & 36.8  & 20.3  &  /     &  \\
    mPLUG-OWL2         & 48.8  & 72.7  & 33.3  &  /     &  \\
    InternLM-XC        & 51.6  & 72.9  & 39.7  &  /     &  \\
    LHRS-Bot           & 91.3  & 93.2  & 51.4  &  /     &  \\
    \midrule
    GeoChat            & 72.0  & 86.5  & 43.4  & 46.6  & \makecell{GeoChat \\ \citep{GeoChat}} \\
    \midrule
    SkySenseGPT        & 92.3  & 97.0 & /      & 48.0  & \makecell{SkySenseGPT\\ \citep{SkySenseGPT}}\\
    \midrule
    H\textsuperscript{2}RSVLM  & 89.3  & 97.0 &   /    &   /    & \makecell{VHM \\ \citep{pang2025vhm}}  \\
    \midrule
    \multicolumn{6}{l}{\footnotesize \textit{\textcolor{darkgray}{Closed-source Commercial Vision-Language Models}}} \\
    GPT-4o                  & \textbf{74.7}     & 90.2     & \textbf{56.8}  & \textbf{68.1} &  \multirow{17}{*}{this study}  \\

    \multicolumn{6}{l}{\footnotesize \textcolor{darkgray}{\textit{Open-source Vision-Language Models}}}  \\
    Qwen-VL-Chat    & 55.5     & 73.5     & 26.2            & 46.8  &  \\
    Qwen2-VL         & 65.3     & 87.8     & 36.1            & 63.6  &  \\
    LLaVA-1.5       & 37.2     & 44.4     & 31.1            & 55.0  &  \\
    LLaVA-1.6       & 48.9     & 69.1     & 43.4            & 64.6 &  \\
    TinyLLaVA       & 54.8     & 76.9     & 42.9            & 62.1  &  \\
    CogVLM\_chat    & 53.7     & 76.5     & 51.6            & 56.5  &  \\
    Qwen3-VL         & 69.5     & 85.7    & 50.9            & 67.5    \\
    InternVL3.5     & 72.5     & \textbf{92.4}  & 53.2            & 60.0    \\
    GLM-4.6V-Flash  & 72.8     & \textbf{92.4}  & 28.5            & 66.0    \\

    \multicolumn{6}{l}{\footnotesize \textcolor{darkgray}{ \textit{Open-source Remote Sensing Vision-Language Models}}} \\
    GeoChat         & 68.7     & 90.3     & 35.4        & 41.1  &  \\
    RS-LLaVA        & 66.8     & 81.5     & 50.4        & 57.7  &  \\
    Geo-R1          & 64.1     & 82.1     & 33.8        & 56.1    \\
    Falcon          & 40.2     & 54.6     & 17.0        & 55.6     \\
    ZoomEarth       & 53.8     & 74.5     & 19.1        & 36.2     \\
    GeoLLaVA-8K     & 4.6      & 3.6      & 0.0         & 31.7      \\
    \bottomrule 
    \end{tabular*}
  \label{tab:sceneC}
\end{table}

\subsection{Comments on RS-MLLMs}
\textbf{Limited standardized evaluation.} Table~\ref{tab:evaluation} summarizes the quantitative evaluation protocols and datasets used in representative RS-MLLMs. Existing studies have covered diverse tasks, including scene classification, visual question answering, visual grounding, image captioning, object detection, segmentation, region-level captioning, and change detection. However, their evaluation settings remain inconsistent. Many models are evaluated on held-out splits of constructed instruction datasets or benchmarks closely related to their training data, while independent ZS or out-of-domain evaluations are still limited. This makes it difficult to distinguish whether the reported gains arise from general multimodal reasoning, RS-specific knowledge, or dataset/task overlap. Moreover, complex abilities such as multi-turn interaction, grounded description, active perception, and high-resolution reasoning are often demonstrated qualitatively but lack standardized quantitative protocols. These limitations motivate a unified evaluation of RS-MLLMs and general CV-MLLMs.

\textbf{Template-dependent task execution.} Many RS-MLLMs rely on task-specific identifiers or predefined templates, such as ``grounding'', ``identify'', ``refer'', or ``[detection]'', to trigger specific outputs. Although such designs can improve task execution under predefined settings, they may also make model behavior sensitive to prompt formats. Recent models have started to mitigate this issue. For example, Falcon introduces dynamic prompt training by sampling semantically equivalent instructions from a predefined prompt pool, aiming to reduce reliance on task-specific tokens~\citep{yao2025falcon}. Nevertheless, its training still depends on standardized task templates and structured output formats. This suggests that current RS-MLLMs have not fully moved beyond template-defined task execution, which may limit flexible instruction following in open-ended RSISU scenarios.

\begin{table}[!t]
  \centering
  \scriptsize
  \setlength{\tabcolsep}{0.1pt}
  \renewcommand{\arraystretch}{1.2}
  \caption{Comparative VQA performance of RS-MLLMs and CV-MLLMs on the test set of RSVQA-HR. Bold numbers denote the best performance in this study.}
    \begin{tabular*}{\hsize}{@{\extracolsep{\fill}}lcccc}
    \toprule
     \textbf{Model} & \textbf{Presence} & \textbf{Comparison} & \textbf{Avg} & \textbf{Source} \\
    \midrule
    Qwen-VL        & 66.4  & 60.4  & 63.1  & \multirow{4}[1]{*}{\makecell{GeoChat \\ \citep{GeoChat}}}  \\
    LLaVA-1.5      & 69.8  & 67.3  & 68.4  &  \\
    MiniGPTv2      & 40.8  & 50.9  & 46.5  &  \\
    GeoChat        & 58.5  & 83.2  & 72.3 &  \\
    \cmidrule{1-5}    
    Qwen-VL-Chat   & 69.8  & 67.3 & 68.4  & \multirow{3}[1]{*}{\makecell{EarthGPT \\ \citep{EarthGPT}}} \\
    Sphinx         & 64.3  & 74.8 & 69.8  &  \\
    EarthGPT       & 62.8  & 79.5 & 72.1  &  \\
    \cmidrule{1-5} 
    SkySenseGPT               & 69.1  & 84.1  & 76.6  & \makecell{SkySenseGPT \\ \citep{SkySenseGPT}} \\
    \cmidrule{1-5}   
    H\textsuperscript{2}RSVLM & 65.0  & 83.7  & 74.4  & \makecell{VHM \\ \citep{pang2025vhm}} \\
    \cmidrule{1-5} 
    RS-LLaVA                  & 70.4  & 78.7  & 75.0  &  \makecell{RS-LLaVA \\  \citep{RS-LLaVA}} \\
    \cmidrule{1-5}
    \multicolumn{5}{l}{\footnotesize \textit{\textcolor{darkgray}{Closed-source Commercial Vision-Language Models}}} \\
    GPT-4o                    & 68.4     & 67.4     & 67.9     & \multirow{15}[1]{*}{this study}  \\
    
    \multicolumn{5}{l}{\footnotesize \textit{\textcolor{darkgray}{Open-source Vision-Language Models}}} \\
    Qwen-VL-Chat    & 62.9     & 53.4     & 58.1  &  \\
    Qwen2-VL         & 65.9     & 77.8     & 71.9 &  \\
    LLaVA-1.5       & 64.4     & 39.0     & 51.7  &  \\
    LLaVA-1.6       & 65.2     & 68.6     & 66.9  &  \\
    TinyLLaVA       & 71.0     & 69.9     & 70.5  &  \\
    CogVLM\_chat    & 59.3     & 68.9     & 64.1     &  \\
    Qwen3-VL         & 63.8     & 76.3     & 70.1   &    \\
    InternVL3.5     & 63.6     & 83.0     & 73.3    &   \\

    \multicolumn{5}{l}{\footnotesize \textcolor{darkgray}{ \textit{Open-source Remote Sensing Vision-Language Models}}} \\
    GeoChat         & 57.2    & 80.1     & 68.7  &  \\
    RS-LLaVA        & 74.1    & 80.7     & 77.4        &  \\
    Geo-R1          & 61.6    & 74.4     & 68.0    &   \\
    Falcon          & \textbf{75.8} & \textbf{86.6} & \textbf{81.2}    &      \\
    ZoomEarth       & 54.2    & 63.5    & 58.8    &      \\
    GeoLLaVA-8K     & 66.7    & 66.2    & 67.4    &         \\
    \bottomrule
    \end{tabular*}%
  \label{tab:rsvqa}%
\end{table}%

\begin{table*}[t]
  \centering
  \scriptsize
  \renewcommand{\arraystretch}{1.2}
  \caption{Comparative performance of RS-MLLMs and CV-MLLMs on the LHRS-Bench dataset. Bold numbers denote the best performance in our study. The top results are from \citep{LHRS-Bot}, while the rest are tested in \textbf{this study}. ID: Identity; Col: Color; Orient: Orientation; Shp: Shape; Area: Area; Res: Resolution; Mod: Modality; Loc: Location; Dist: Distance; Qty: Quantity; Reas: Reason; Avg: Average accuracy.}
    \begin{tabular*}{\hsize}{@{\extracolsep{\fill}}lcccccccccccc}
    \toprule
      \textbf{Method} & \textbf{ID} & \textbf{Col} & \textbf{Orient} & \textbf{Shp} & \textbf{Area} & \textbf{Res} & \textbf{Mod} & \textbf{Loc} & \textbf{Dist} & \textbf{Qty} & \textbf{Reas} & \textbf{Avg} \\
    \midrule
    LLaVA-1.5      & 42.9  & 31.0  & 15.4  & 43.2  & 44.0  & 23.8  & 21.7  & 35.8  & 27.3  & 21.9  & 54.4  & 37.5  \\
    MiniGPTv2      & 20.2  & 15.9  & 2.6   & 8.1   & 13.3  & 14.3  & 4.4   & 15.2  & 13.6  & 9.5   & 37.0  & 16.9   \\
    InstructBLIP   & 20.4  & 22.1  & 5.1   & 24.3  & 22.7  & 38.1  & 0.0   & 13.7  & 22.9  & 13.1  & 30.4  & 18.9   \\
    mPLUG-OWL2     & 41.5  & 45.1  & 12.8  & 40.5  & 32.0  & 33.3  & 0.0   & 30.9  & 13.6  & 32.1  & 54.3  & 37.0   \\
    Qwen-VL-Chat   & 31.4  & 32.7  & 7.7   & 32.4  & 29.3  & 19.1  & 4.4   & 24.0  & 21.5  & 21.2  & 56.5  & 28.7    \\
    InternLM-XComposer & 34.4  & 31.0  & 7.7   & 37.8  & 26.7  & 0.0   & 8.7   & 33.3  & 22.7  & 18.3  & 65.2  & 31.1    \\
    LHRS-Bot       & 44.6  & 42.5  & 23.1  & 67.6  & 25.3  & 42.9  & 32.4  & 56.5  & 22.7  & 18.3  & 73.9  & 39.4   \\
    \midrule
    \multicolumn{13}{l}{\footnotesize  \textcolor{darkgray}{\textit{Closed-source Vision-Language Models}}} \\
    GPT-4o          & 70.5 & 68.1 & 43.6 & 75.7 & 68.0 & \textbf{76.2} & 65.2 & 68.6 & 72.7 & 50.4 & 87.0 & 67.8  \\
    
    \multicolumn{13}{l}{\footnotesize  \textcolor{darkgray}{\textit{Open-source Vision-Language Models}}}  \\
    Qwen-VL-Chat    & 45.4  & 43.4  & 28.2  & 48.7  & 42.7  & 28.6  & 21.7  & 44.6  & 27.3  & 34.3  & 60.9  & 38.7  \\
    Qwen2-VL         & 68.0 & 66.4 & 35.9  & 70.3 & 72.0 & 52.4  & 47.8 & 70.1 & 59.1 & 50.4 & 80.4 & 61.2 \\
    LLaVA-1.5       & 39.8  & 37.2  & 28.2  & 43.2  & 32.0  & 33.3  & 43.5  & 36.3  & 31.8  & 26.3  & 54.4  & 36.9  \\
    LLaVA-1.6       & 49.1  & 43.4  & 23.1  & 56.8  & 50.7  & 38.1  & 43.5  & 53.4  & 27.3  & 40.9  & 60.9  & 44.3  \\
    TinyLLaVA       & 59.5  & 54.9  & 30.8  & 64.9  & 48.0  & 57.1  & 30.4  & 59.3  & 40.9  & 44.5  & 78.3  & 51.7  \\
    CogVLM\_chat    & 49.4  & 45.1  & 25.6  & 46.0  & 52.0  & 14.3  & 13.0  & 42.7  & 45.5  & 38.7  & 76.1  & 40.8   \\
    Qwen3-VL        & 75.2  &  \textbf{82.3} & 41.0 &  81.1 &  68.0 &  47.6 &  26.1 &  78.9 &  63.6 &  \textbf{58.4} &  \textbf{82.6} & 64.1 \\
    InternVL3.5     & 68.6 &  66.4 &  30.8 &  64.9 &  66.7 &  14.3 &  21.7 &  76.0 &  59.1 &  49.6 &  71.7 & 53.6 \\
    GLM-4.6V-Flash  &  \textbf{77.0} & 77.9  & 38.5  & \textbf{86.5} & 72.0  &  33.3 &  56.5 &  \textbf{86.8} &  \textbf{68.2} &  48.9 &  \textbf{82.6} & \textbf{66.2} \\
    \multicolumn{13}{l}{\footnotesize   \textcolor{darkgray}{\textit{Open-source remote sensing Vision-Language Models}}} \\
    GeoChat         & 43.4  & 38.1  & 23.1  & 46.0  & 33.3  & 23.8  & 30.4  & 40.7  & 22.7  & 36.5  & 71.7  & 37.2  \\
    RS-LLaVA        & 50.2  & 43.4  & 28.2  & 62.2  & 38.7  & 42.9  & 21.7  & 44.1  & 45.5  & 35.0  & 56.5  & 42.6   \\
    Geo-R1          &  71.9 &  70.8 &  \textbf{56.4} &  73.0 &  \textbf{73.3} &   4.8 &  26.1 &  75.0 &  54.5 &  49.6 &  76.1 & 57.4 \\
    ZoomEarth       &  71.5 &  81.4 &  38.5 &  73.0 &  72.0 &  47.6 &  52.2 &  75.0 &  \textbf{68.2} & 51.8 & 73.9 & 64.1 \\
    GeoLLaVA-8K     &  36.4 &  24.8 &  23.1 &  62.2 &  29.3 &   9.5 &  \textbf{69.6} &  39.2 &  27.3 &  32.1 &  32.6 & 35.1 \\
    \bottomrule
    \end{tabular*}%
  \label{tab:LHRS-Bench}%
\end{table*}%

\section{Evaluation of MLLMs for RSISU}
\label{sec:performance}

\textbf{MLLMs' capacities and enabled tasks.}
The capabilities of MLLMs can be classified in various ways. One approach involves a hierarchical taxonomy of human-like capabilities, with perception and reasoning as the top-level categories. Perception is further divided into image-wise, instance-wise, and pixel-wise capabilities, while reasoning encompasses logical and attribute-based reasoning \citep{zhou2024visionlanguagegeofoundationmodelsurvey}.

Alternatively, MLLMs can be categorized based on the format of inputs and outputs, including languages, numbers, coordinates, and images. For answers that consist solely of text, preliminary tasks include VQA and captioning, which can be further classified into scene-level, region-level (focusing on specific areas indicated by text or bounding boxes, such as in referring expression generation), sensor-level (e.g., SAR-level for SAR images), and language-level (e.g., supporting languages like Chinese in addition to English). Questions can also be categorized into types such as counting, geospatial localization, or physical property assessment. For tasks where text alone is insufficient, grounding tasks require bounding boxes in the answers. Visual grounding  is divided into two specific tasks: phrase localization, which involves identifying and annotating all objects mentioned in a sentence and referring expression comprehension (REC), which focuses on locating a single object indicated by a language description. Additional tasks that fall into this category include grounded IC, grounded question answering, referring expression segmentation, open-vocabulary detection, and open-vocabulary segmentation.
 
This section will provide a balanced comparison between RS-MLLMs and representative CV-MLLMs (as listed in Table \ref{tab:cvModels}) on common RSISU tasks. \textbf {Although results reported in existing literature are included as references, our discussion primarily relies on the results obtained in this study, as indicated in the tables, to ensure consistency in experimental settings and evaluation protocols.} The accuracies reported in this study may differ from those found in the literature, as observed for models like LLaVA-1.5 and GeoChat in Table \ref{tab:sceneC}. These discrepancies can be attributed to variations in the evaluation methodologies and differences in the prompts used. For example, some models include a period at the end of an answer, which can affect accuracy when answers are directly compared to ground truth. Therefore, post-processing is necessary to achieve more reasonable accuracy. Additionally, the inherent ambiguity of language presents challenges in standardizing evaluations across different implementations, as MLLM performance can be influenced by prompt selection. This variability complicates the effort to make strictly fair comparisons between different MLLMs, as they may respond better to specific prompts.

\begin{table}[t]
    \centering
    \scriptsize
    \setlength{\tabcolsep}{0.1pt}
    \renewcommand{\arraystretch}{1.2}
    \caption{Visual grounding performance of RS-MLLMs and CV-MLLMs on RSVG, DIOR-RSVG, and VRSBench dataset. The $\dagger$ symbol denotes metrics evaluated in this study, whereas remaining results are adopted from RSThinker~\citep{RSThinker}.}
    \begin{tabular*}{\hsize}{@{\extracolsep{\fill}}lccccccccc}
    \toprule
    \textbf{Method}    & \multicolumn{3}{c}{\textbf{VRSBench}} & \multicolumn{3}{c}{\textbf{DIOR-RSVG}} & \multicolumn{3}{c}{\textbf{RSVG} }    \\ \cmidrule{2-4} \cmidrule{5-7} \cmidrule{8-10}  
                        & @0.5    & @0.7  & mIoU   & @0.5    & @0.7  & mIoU  & @0.5    & @0.7  & mIoU  \\ 
     \midrule
    \multicolumn{10}{l}{\footnotesize \textit{\textcolor{darkgray}{Closed-source Commercial Vision-Language Models}}} \\
    Claude-sonnet-4     & 11.1 & 2.4 & 16.66     & 17.6 & 1.2 & 25.33    & 24.0 & 7.0 & 24.99 \\
    Gemini-2.0-flash    & 22.9 & 6.3 & 28.59     & 20.8 & 3.3 & 27.45    & 19.5 & 4.5 & 24.07 \\
    ChatGPT-5           & 14.4 & 2.3 & 22.71     & 26.1 & 3.3 & 28.37     & 18.5 & 3.5 & 20.59 \\

    \multicolumn{10}{l}{\footnotesize \textit{\textcolor{darkgray}{Open-source Vision-Language Models}}} \\
    MiniGPT-v2                 & 32.1 & 16.3 & 33.96    & 29.4 & 10.2 & 29.43     & 12.0 & 3.0 & 15.65 \\
    Qwen2.5VL                & 45.2 & 20.6 & 42.45    & 36.3 & 15.9 & 34.34     & 1.0  & 0.0 & 7.24 \\
    Qwen3-VL$^\dagger$        & 53.3 & 32.4 & 47.75    & 57.8 & 44.1 & 45.72     & 23.0 & 11.0 & 24.48\\
    InternVL3.5$^\dagger$     & 27.2 & 9.6  & 28.49    & 26.6 & 8.2  & 26.89     & 16.7 & 4.5 & 17.75\\

    \multicolumn{10}{l}{\footnotesize  \textit{\textcolor{darkgray}{Open-source Reasoning Vision-Language Models}}} \\
    GLM-4.1V-Thinking    & 63.8 & 47.0 & 60.69    & 59.6 & 43.7 & 57.41  & 43.0 & 30.5 & 42.27 \\
    GLM-4.6V-Flash$^\dagger$   & 47.6 & 25.0 &  40.74   & 52.0 & 39.7 & 47.72    & 23.1 & 11.3 & 22.19\\

    \multicolumn{10}{l}{\footnotesize \textit{\textcolor{darkgray}{Open-source Remote Sensing Vision-Language Models}}} \\
    GeoChat              & 56.3 & 24.6  & 53.50       & 31.4   & 11.0  & 34.99        & 5.5 & 0.5 & 12.55 \\ 
    VHM                  & 33.9 & 10.0  & 34.91       & 55.9   & 35.5  & 49.90        & 2.5 & 0.0 & 5.80  \\
    SkySenseGPT          & 63.4 & 26.0  & 54.60       & 60.8  & 26.5  & 53.18        & 39.5 & 17.5 & 38.54 \\
    EarthDial            & 14.1 & 7.8   & 13.04       & 46.1   & 30.2  & 39.46        & 42.0 & 24.0 & 38.49 \\
    RSThinker         & \textbf{90.4} & \textbf{77.2} & \textbf{80.79}    & \textbf{93.1} & \textbf{90.2}  & \textbf{89.02} & \textbf{64.0} & \textbf{54.5} & \textbf{59.74} \\ 
    Geo-R1$^\dagger$     & 26.1 & 10.2  & 28.81       & 35.6   & 16.9  & 26.57        & 6.4  & 1.2 & 10.46\\
    ZoomEarth$^\dagger$  & 22.5 & 11.7  & 23.13       & 34.6   & 21.9  & 23.65        & 6.2  & 1.3 & 9.48\\
    
    \bottomrule
    \end{tabular*}
  \label{tab:vg}
\end{table}

\subsection{Evaluation Protocol}
We adopt a standardized evaluation protocol to ensure fair and reproducible comparisons across all models. All methods are evaluated using publicly released checkpoints or API versions under a zero-shot setting, without any task-specific fine-tuning or adaptation.

To account for potential benchmark contamination, Table~\ref{tab:contamination_audit} summarizes benchmark release dates, 
possible training-data overlap, evaluation settings, and contamination risks. Accordingly, we distinguish strict zero-shot evaluation from no-adaptation inference, benchmark-aware evaluation, and transfer learning. The reported risk levels indicate potential exposure rather than confirmed contamination.

For visual inputs, we follow each model's native preprocessing pipeline to avoid introducing additional bias. All models are evaluated under consistent task-specific prompt templates with strict output format constraints (e.g., binary answers or multiple-choice letters), ensuring identical instruction conditions across methods.

For generation, we adopt greedy decoding ($\text{do\_sample}=\texttt{False}$) for all models with task-specific maximum token lengths. Model outputs are post-processed using a unified deterministic parsing pipeline, including reasoning-tag removal and task-specific normalization for classification, VQA, multiple-choice, and visual grounding tasks.

We evaluate five representative tasks, including scene classification, visual question answering, multi-label recognition, multiple-choice reasoning, and visual grounding, across eleven benchmark datasets. Classification tasks are evaluated using Overall Accuracy, while VQA and multiple-choice tasks report accuracy. Visual grounding is evaluated using mean Intersection-over-Union (mIoU) and Acc@0.5/0.7.

\subsection{Performance on task-specific RS datasets}
\subsubsection{Scene Classification}
Table \ref{tab:sceneC} presents the comparative remote sensing image scene classificaition results for both single-label and multi-label tasks, which are commonly evaluated in current RS-MLLM studies. On relatively simple datasets, such as AID~\citep{AID}, WHU-RS19~\citep{WHU-RS19} and EuroSAT~\citep{Eurosat}, several literature-reported RS-MLLMs achieve highly competitive performance. In particular, LHRS-Bot, SkySenseGPT, and H\textsuperscript{2}RSVLM obtain the best or near-best results, even surpassing the proprietary GPT-4o on these two benchmarks. This demonstrates the effectiveness of remote-sensing-specific fine-tuning for scene-level recognition. However, this advantage becomes less evident under our unified evaluation setting. On more challenging benchmarks, including EuroSAT with 10 m GSD imagery and AID-multi with multi-label annotations, CV-MLLMs, including both open-source and closed-source models, generally outperform existing RS-MLLMs. These results suggest that current RS-MLLMs still face limitations when handling lower-resolution imagery and complex multi-label scene understanding.
\begin{table}[h]  
  \centering 
  \scriptsize
  \setlength{\tabcolsep}{0.1pt}
  \renewcommand{\arraystretch}{1.2}
  \caption{Comparative performance of RS-MLLMs and CV-MLLMs on the FIT-RSRC dataset. Bold numbers indicate the best performance among this study (Sub: subject, Obj: Object, Rel: Relation, Exi: Existence, TL: transfer learning, Acc: accuracy). The $\ast$ symbol denotes metrics are evaluated on transfer learning.}  
  \begin{tabular*}{\hsize}{@{\extracolsep{\fill}}lcccccc}  
    \toprule
    \textbf{Method} & \textbf{Sub} & \textbf{Obj} & \textbf{Rel} & \textbf{Exi} & \textbf{Acc} & \textbf{Source} \\
    \midrule  
    LLaVA-1.5-7B   & 6.4   & 11.2  & 21.0  & 7.8  & 11.6 &  \multirow{5}{*}{\makecell{SkySenseGPT \\ \citep{SkySenseGPT}}} \\
    TinyLLaVA      & 22.0  & 18.0  & 12.0  & 55.6 & 26.9 &   \\
    LLaVA-HR-7B    & 7.8   & 13.4  & 32.6  & 47.8 & 25.4 &   \\
    GeoChat        & 7.6   & 8.6   & 25.6  & 45.6 & 21.9 &   \\
    SkySenseGPT$^{\ast}$    & 25.2  & 54.0 & 50.4 & 92.2 & 55.5 & \\
    \midrule
    \multicolumn{7}{l}{\footnotesize \textit{\textcolor{darkgray}{Closed-source Vision-Language Models}}} \\
    GPT-4o         & 56.2 & 36.7  & 49.8  & 80.5 & 55.8 & \multirow{18}{*}{this study} \\

    \multicolumn{7}{l}{\footnotesize  \textit{\textcolor{darkgray}{Open-source Reasoning Vision-Language Models}}} \\
    Qwen-VL-Chat   & 30.2    & 28.6  & 22.8  & 49.3   & 32.7 &   \\
    LLaVA-1.5      & 33.0    & 20.1  & 29.9  & 58.2   & 35.3 &   \\
    LLaVA-1.6      & 54.2    & 37.9  & 39.0  &  62.0  & 48.3 &   \\
    TinyLLaVA      & 41.6    & 27.9  & 28.7  & 69.2   & 41.8 &   \\
    CogVLM\_chat   & 42.8    & 45.1  & 37.3  & 51.6  & 44.2 &   \\    
    Qwen3-VL        & 63.2    & \textbf{54.3} & \textbf{52.7}  & \textbf{88.3} & \textbf{64.3}    &     \\
    InternVL3.5    & 51.6    & 16.3      & 38.0      & 74.7   & 44.9      &     \\
    GLM-4.6V-Flash & \textbf{86.6} & 43.8      & 49.3      & 63.9    &   60.4     &      \\
    
    \multicolumn{7}{l}{\footnotesize \textit{\textcolor{darkgray}{Open-source  Remote Sensing Vision-Language Models}}} \\
    GeoChat        & 44.7   & 32.6       & 27.5        & 43.2    & 37.0     &       \\
    RS-LLaVA       & 46.0   & 26.9        & 43.0        & 54.8    & 42.7    &       \\
    Falcon         & 7.3     & 6.4       & 12.3        & 6.2     & 8.2      &       \\
    Geo-R1         & 59.9    & 41.2      & 45.1       & 67.8     & 53.3     &       \\
    ZoomEarth      & 46.5    & 31.2      & 34.2        & 60.6    & 42.9     &       \\
    GeoLLaVA-8K    & 25.5    & 11.5      & 33.9       & 35.4     & 26.7     &       \\
    \bottomrule  
  \end{tabular*}  
  \label{tab:fitrc}  
\end{table}

The results also reveal clear differences in dataset difficulty. WHU-RS19 appears to be the least challenging benchmark, partly because it contains only 19 scene categories and consists of very high-resolution images. AID is also composed of VHR images but includes a larger number of categories, making it relatively more difficult. In contrast, EuroSAT is based on Sentinel-2 imagery with a 10 m GSD, which differs substantially from both natural images commonly used to train general CV-MLLMs and the VHR imagery frequently adopted in RS-MLLM training. This domain and resolution gap makes EuroSAT particularly challenging for existing RS-oriented models.

Overall, the findings highlight the strong generalization ability of recent MLLMs in the remote sensing domain. Among the closed-source models, GPT-4o achieves the best overall performance, especially on EuroSAT and AID-multi. Nevertheless, it is not consistently the best model across all datasets, as InternVL3.5 and GLM-4.6V-Flash achieve higher accuracy on WHU-RS19 under our evaluation. Among the open-source CV-MLLMs evaluated in this study, InternVL3.5, GLM-4.6V-Flash, and Qwen3-VL show the most competitive performance, clearly outperforming earlier models such as Qwen2-VL, LLaVA-1.6, Tiny-LLaVA, CogVLM\_chat, and Qwen-VL-Chat on most benchmarks. In contrast, GeoLLaVA-8K performs poorly across most datasets, suggesting that its SFT process may lead to limited generalization, with the model overfitting specific instruction patterns and failing to robustly answer diverse scene classification queries.

\subsubsection{Visual Question Answering}
Table \ref{tab:rsvqa} presents the VQA performance of RS-MLLMs and CV-MLLMs on Test Set 2 of RSVQA-HR~\citep{RSVQA}, a widely used benchmark in the literature. Following common evaluation practice, area and count questions are excluded. For RSVQA-HR, we follow a controlled evaluation protocol by selecting two representative question types, i.e., Presence and Comparison, which are the most commonly studied reasoning categories in prior RS-VQA literature and directly reflect object existence reasoning and relational reasoning ability in remote sensing images. In this study, the results are obtained from a random sample of 528 Presence-type and 528 Comparison-type questions. Overall, RS-MLLMs show clear advantages over general CV-MLLMs on this benchmark, especially on the more challenging Comparison-type questions. Among them, Falcon achieves the best overall performance, while RS-LLaVA also obtains strong results. Literature-reported models, such as SkySenseGPT and H\textsuperscript{2}RSVLM, also show competitive performance, further indicating the effectiveness of remote-sensing-specific instruction tuning for RSVQA-HR.

Among the eight CV-MLLMs evaluated in this study, InternVL3.5 achieves the highest average accuracy, followed by Qwen2-VL, TinyLLaVA, and Qwen3-VL. Although GPT-4o shows stable performance across the two question types, it does not achieve the best results on this task and is surpassed by several open-source CV-MLLMs. These results suggest that recent open-source general-purpose MLLMs have achieved improved transferability to remote sensing VQA, but they still lag behind the strongest RS-oriented models. The larger performance gaps on Comparison questions suggest that fine-grained object comparison and spatial reasoning remain key challenges for general-purpose MLLMs in high-resolution remote sensing imagery.
\begin{figure*}[t]
	\centering
	\includegraphics[width=0.98\textwidth]{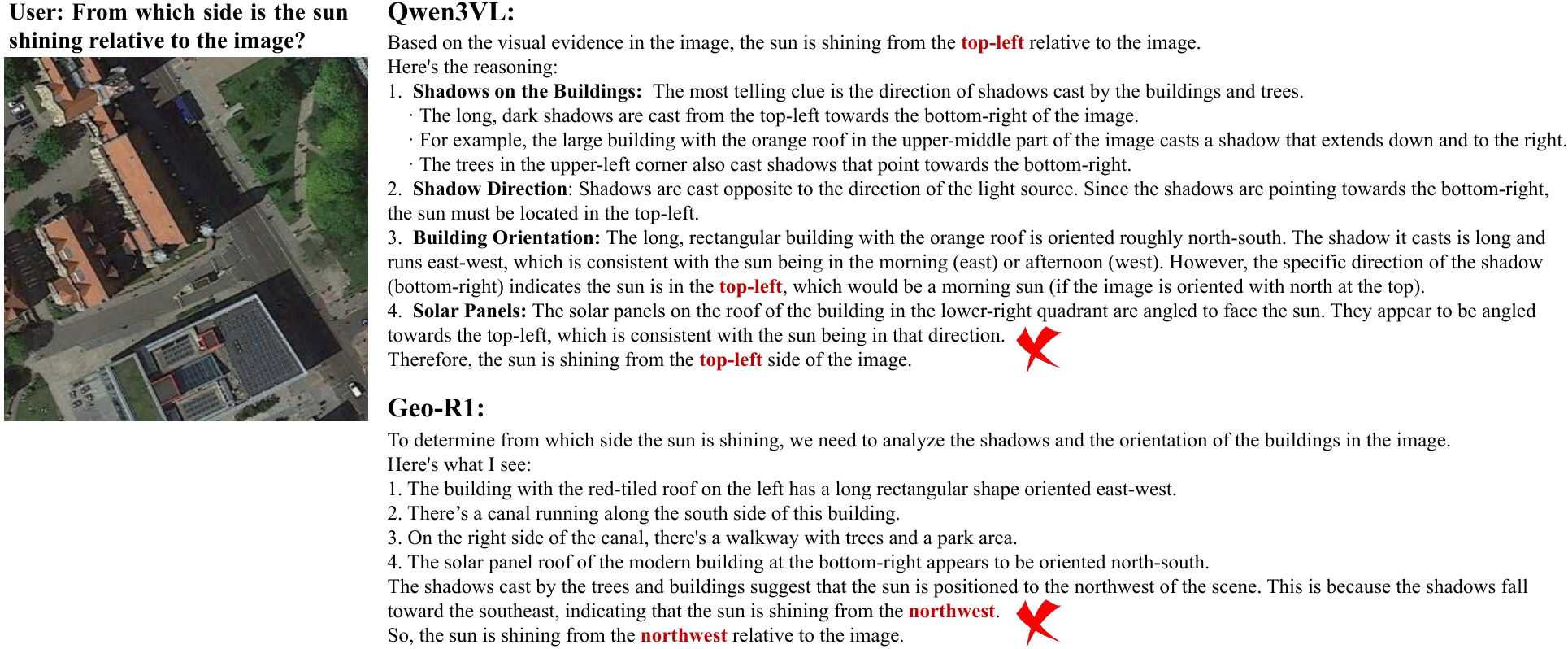}
	\caption{Complex RSISU reasoning cases from  Qwen3-VL and Geo-R1.
	}
	\label{fig:complex-sun}
\end{figure*}
\subsubsection{Visual Grounding}

Table \ref{tab:vg} presents the visual grounding results on three representative remote sensing grounding benchmarks, including VRSBench~\citep{VRSBench}, DIOR-RSVG~\citep{DIOR-RSVG}, and RSVG~\citep{SunRSVG}. Following common grounding evaluation protocols, model performance is measured using Acc@0.5, Acc@0.7, and mIoU, where Acc@0.5 and Acc@0.7 indicate whether the predicted bounding box overlaps with the ground-truth box by more than 0.5 and 0.7 Intersection-over-Union (IoU), respectively.

Overall, RS-MLLMs exhibit stronger grounding performance than general CV-MLLMs, especially on VRSBench and DIOR-RSVG. This advantage may be partly attributed to remote-sensing-specific instruction tuning and the possible inclusion of related grounding datasets, such as RSVG or DIOR-RSVG, in the training data of some RS-oriented models. Among all evaluated models, RSThinker achieves the best performance across all three datasets and all evaluation metrics, demonstrating a clear advantage in precise object localization. In particular, its substantial gains under the stricter Acc@0.7 metric suggest that it can produce more accurate bounding boxes rather than merely identifying approximate target regions.

\begin{figure*}[t]
	\centering
	\includegraphics[width=\textwidth]{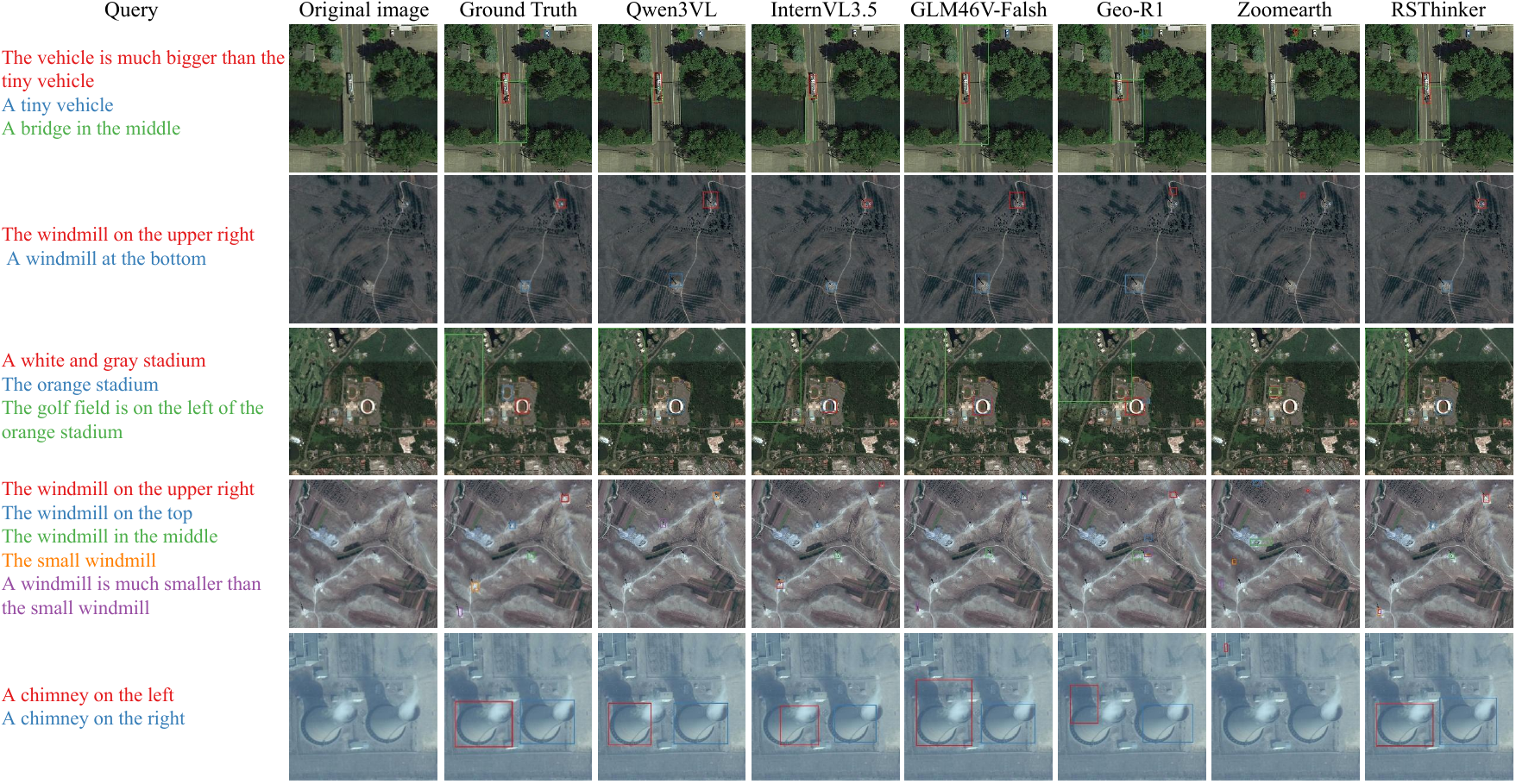}
	\caption{Visual grounding performance of representative MLLMs on the DIOR-RSVG dataset. Different colors are utilized to distinguish the queries. }
	\label{fig:vg}
\end{figure*}

In addition, reasoning-oriented VLMs, such as GLM-4.1V-Thinking, also show competitive performance, indicating that explicit reasoning ability is beneficial for visual grounding tasks that require understanding referring expressions, spatial relationships, and object locations. In contrast, closed-source commercial VLMs and most general CV-MLLMs show limited localization precision, especially under the Acc@0.7 metric. Moreover, Figure~\ref{fig:vg} further illustrates the promising performance of typical MLLMs. These results suggest that fine-grained visual grounding in remote sensing remains a challenging task for current general-purpose MLLMs.

\begin{figure}[t]
\centering
\includegraphics[width=0.48\textwidth]{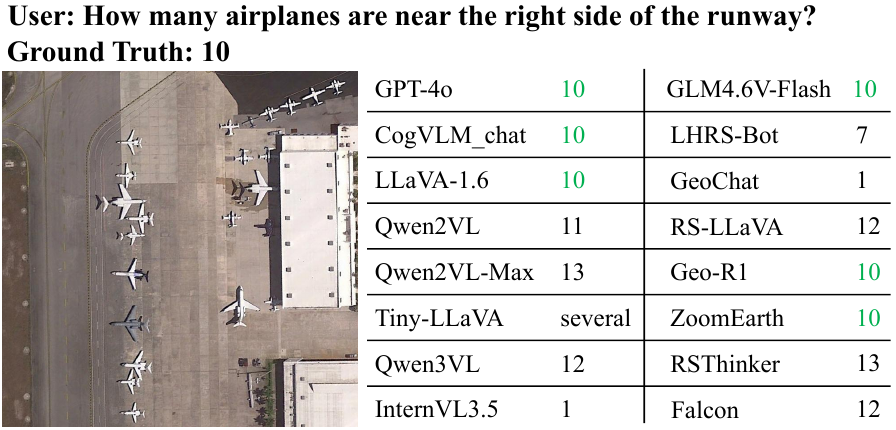}
\caption{Representative example of counting and spatial reasoning by state-of-the-art MLLMs.}
\label{fig:counting}
\end{figure}

\subsection{Performance on MLLM-specific RS Benchmarks}
\subsubsection{LHRS-Bench Benchmark}
LHRS-Bench is a comprehensive benchmark designed for evaluating RS-MLLMs. It contains 690 single-choice questions organized into five top-level dimensions: Recognition, Imagery, Spatial-awareness, Quantity, and Reasoning. Recognition focuses on identifying geographical objects and their attributes, such as color and shape. Imagery evaluates the understanding of image resolution and modality. Spatial-awareness assesses the ability to reason about object locations and distances, while Quantity and Reasoning evaluate counting ability and domain-specific visual reasoning. These dimensions are further divided into 11 sub-dimensions. To reduce the risk of data leakage, LHRS-Bench uses 108 images collected exclusively from Google Earth rather than public remote sensing datasets. Its single-choice format, with 2--4 candidate answers for each question, enables quantitative and objective evaluation.

Table \ref{tab:LHRS-Bench} presents the evaluation results on LHRS-Bench. Overall, recent general-purpose MLLMs demonstrate strong performance on this benchmark. GPT-4o achieves the best average accuracy among all evaluated models. Among open-source CV-MLLMs, GLM-4.6V-Flash obtains the highest average accuracy, followed by Qwen3-VL and Qwen2-VL. These models substantially outperform early CV-MLLMs, such as MiniGPTv2, InstructBLIP, and LLaVA-1.5, indicating the rapid progress of general visual-language models and their strong transferability to remote sensing understanding.

In contrast, existing RS-MLLMs do not consistently outperform general CV-MLLMs on LHRS-Bench. Although ZoomEarth and Geo-R1 achieve competitive performance, several RS-oriented models, such as GeoChat, RS-LLaVA, and GeoLLaVA-8K, remain clearly behind recent general-purpose MLLMs. This suggests that remote-sensing-specific instruction tuning alone is insufficient to guarantee strong generalization across diverse question types. The category-wise results further show clear performance variation across different task types. Most models perform relatively well on Identity, Shape, and Reasoning questions, but still struggle with Orientation, Modality, and Resolution. These results indicate that LHRS-Bench requires not only remote sensing domain knowledge, but also robust visual perception, spatial reasoning, and instruction-following ability.

\subsubsection{FIT-RSRC Benchmark}

The FIT-RSRC dataset is designed to evaluate fine-grained relation comprehension in remote sensing scenes using multiple-choice questions. It contains four question types: Subject, Object, Relation, and Existence, each accounting for 25\% of the dataset. Negative and unanswerable cases are also included to assess model robustness in ambiguous or non-existent relation scenarios. Most questions provide four candidate answers, while a smaller portion provides five options. To improve the quality of distractors, the incorrect options are carefully checked against a curated commonsense word list, preventing models from identifying the answer purely through textual priors.

Table \ref{tab:fitrc} presents the comparative performance of RS-MLLMs and CV-MLLMs on FIT-RSRC. Different from some traditional remote sensing tasks, the strongest results on this benchmark are achieved by recent general-purpose CV-MLLMs. Qwen3-VL obtains the highest average accuracy among all models, followed by GLM-4.6V-Flash. Both models outperform GPT-4o and all evaluated RS-MLLMs under the reported settings. This indicates that FIT-RSRC requires strong fine-grained visual perception, compositional reasoning, and instruction-following ability, rather than relying solely on remote sensing domain-specific tuning.

Among RS-MLLMs, Geo-R1 achieves the best average accuracy in this study, but it still lags behind the strongest general-purpose CV-MLLMs. SkySenseGPT also achieves competitive performance, especially on Object, Relation, and Existence questions. However, its result is obtained under a transfer learning setting, making it not directly comparable with strict zero-shot evaluations. The question-type analysis shows that Existence is generally the easiest category for most models, while Object and Relation remain more challenging because they require fine-grained object recognition and inter-object relationship understanding. Subject questions are handled well by several recent MLLMs, particularly GLM-4.6V-Flash and Qwen-3VL, but earlier models and some RS-MLLMs still show limited performance. These results suggest that current RS-MLLMs still need stronger compositional reasoning and broader instruction generalization to perform well on fine-grained remote sensing relation comprehension.

\subsubsection{XLRS-Bench Benchmark}
XLRS-Bench further complements LHRS-Bench and FIT-RSRC by evaluating MLLMs in ultra-high-resolution remote sensing scenarios. At the L-3 task level, it covers diverse perception and reasoning abilities, including counting, land-use classification, object recognition, multi-scale perception, spatial relationship understanding, anomaly detection, route planning, and change reasoning. In this study, the tested models are evaluated under a zero-shot inference protocol, where no additional fine-tuning is performed on XLRS-Bench. Nevertheless, since some results are collected from public reports or models released after the benchmark became available, they should be interpreted as benchmark-aware evaluations rather than strictly contamination-free zero-shot results. As shown in Table \ref{tab:xlrs_bench_l3_results}, RS-oriented models show strong competitiveness on this benchmark, with ScaleEarth achieving the best average performance, followed by GeoEyes and GeoLLaVA-8K. This differs from LHRS-Bench and FIT-RSRC, where recent general-purpose CV-MLLMs often perform strongly. These results suggest that domain-specific adaptation remains important for ultra-high-resolution remote sensing understanding, especially when models need to integrate fine-grained perception, spatial reasoning, regional semantics, and remote-sensing-specific knowledge.

\subsection{Complex RSISU Cases }
\label{sec:complex}
Figures \ref{fig:complex-sun} and \ref{fig:counting} show two representative failure cases of current MLLMs on complex RSISU tasks requiring spatial reasoning. In Figure~\ref{fig:complex-sun}, the model needs to infer the direction of sunlight based on the shadows cast by buildings. This is relatively straightforward for humans, as the illumination direction can be inferred as opposite to the shadow direction. However, even strong models such as Qwen3-VL and Geo-R1 fail on this case. Qwen3-VL predicts the direction as top-left, while Geo-R1 answers northwest, both of which are incorrect. This suggests that current MLLMs still lack robust shadow-based spatial reasoning ability in remote sensing scenes. In addressing the question in Figure~\ref{fig:counting}, the model needs to understand the runway layout, identify the airplanes, and correctly interpret the spatial term "right". However, more than half of the tested MLLMs fail to provide the correct answer, indicating that fine-grained counting and spatial relation understanding remain challenging for current MLLMs.

\begin{figure}[!t]
	\centering
	\includegraphics[width=0.48\textwidth]{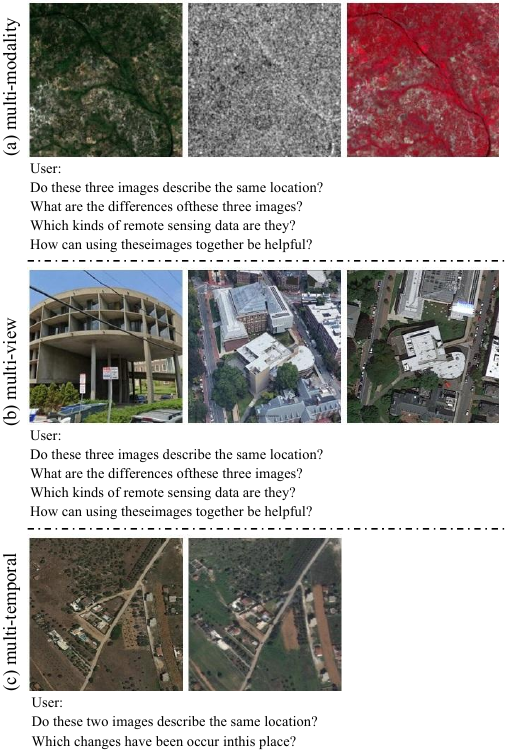}
	\caption{Representative failure cases of current MLLMs on multi-modal, multi-view, and multi-temporal RSISU tasks.}
	\label{fig:multi-views}
\end{figure}

Figure \ref{fig:multi-views} presents representative RSISU examples that require distinguishing among true-color optical images, false-color optical images, and SAR images, as well as performing multi-view image analysis and multi-temporal change detection. When these questions are posed to state-of-the-art MLLMs, the responses are generally unsatisfactory. In particular, most models struggle to handle multiple images as input, which limits their ability to perform cross-view comparison and temporal reasoning. Since such tasks are common in practical RSISU scenarios, these results reveal a substantial gap between the current capabilities of MLLMs and the requirements of real-world remote sensing applications.

\begin{figure}[t]
    \centering
	\includegraphics[width=0.5\textwidth]{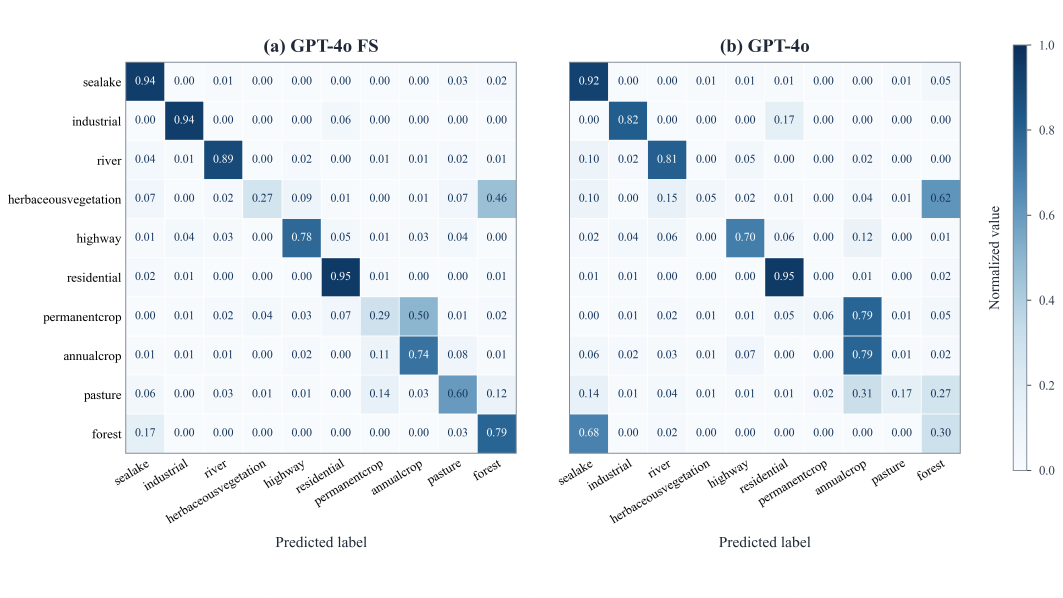}
\caption{Comparison of confusion matrices from the classification on the EuroSAT dataset, showcasing performance before and after applying FS-based prompts on GPT-4o.}
\label{fig:fs_comparison}

\end{figure} 

\begin{table*}[t]
\centering
\caption{Comprehensive VQA results on XLRS-Bench~\citep{xlrs_bench} at the L-3 task level. The best results are highlighted in bold. OC: overall counting; RC: regional counting; OLUC: overall land-use classification; RLUC: regional land-use classification; OCC: object category classification; OCL: object color; OMS: object multi-scale perception; OSR: object spatial relationship; AD: anomaly detection; ECR: embodied commonsense reasoning; RP: route planning; RCCD: regional counting with change detection; CCR: complex change reasoning. Avg denotes average accuracy. The  results are adopted from GeoEyes~\citep{Geoeyes} and ScaleEarth~~\citep{ScaleEarth}.}
\label{tab:xlrs_bench_l3_results}
\setlength{\tabcolsep}{3.0pt}
\renewcommand{\arraystretch}{1.2}
\scriptsize
\resizebox{\textwidth}{!}{
\begin{tabular}{lcccccccccccccc}
\toprule
    \textbf{Model} 
    & \textbf{OC} 
    & \textbf{RC} 
    & \textbf{OLUC} 
    & \textbf{RLUC} 
    & \textbf{OCC} 
    & \textbf{OCL} 
    & \textbf{OMS} 
    & \textbf{OSR} 
    & \textbf{AD} 
    & \textbf{ECR} 
    & \textbf{RP} 
    & \textbf{RCCD} 
    & \textbf{CCR} 
    & \textbf{Avg} \\
    \midrule

    \multicolumn{15}{l}{\footnotesize  \textit{\textcolor{darkgray}{Closed-source Commercial Vision-Language Models}}} \\

    GPT-4o
    & 24.3 & 31.5  & 16.8 & 64.0 & 19.6  & 29.0 & 40.3 & 32.4  & 72.1 & 75.3  &  41.2 & 21.9 & 26.3  & 38.1  \\
    
    Claude 3.7 Sonnet 
    & 27.6 & 22.7 & 17.4 & 68.4 & 30.5 & 29.9 & 63.6 & 27.6 & 64.8 & 78.4 & 34.5 & 27.8 & 32.6 
    & 40.5 \\
    
    GPT-5.2 
    & 30.0 & 37.0 & 17.0 & 70.5 & 43.0 & 41.4 & \textbf{68.3} & 34.0 & 74.0 & 76.0 & 52.0 & 36.7 & 38.0 
    & 47.5 \\

    \multicolumn{15}{l}{\footnotesize   \textit{\textcolor{darkgray}{Open-source Vision-Language Models}}} \\
    Qwen2-VL-7B 
    & 26.7 & 40.0 & 11.0 & 73.0 & 35.9 & 34.6 & 61.7 & 31.8 & 70.0 & 81.0 & 35.0 & 46.7 & 48.0 
    & 45.8 \\
    
    Qwen2.5VL-7B 
    & 33.3 & 40.0 & 31.0 & 77.0 & 40.6 & 40.5 & 66.7 & 36.2 & 68.0 & 72.0 & 27.0 & 38.3 & 45.0 
    & 47.4 \\
    
    Qwen2.5VL-72B 
    & 33.3 & 47.0 & 39.0 & 80.0 & 45.3 & 42.1 & 65.0 & 34.0 & 71.0 & 74.0 & 37.0 & 43.3 & 42.0 
    & 50.2 \\    
    
    InternVL2.5-8B 
    & 38.3 & 37.0 & 10.0 & 77.0 & 33.4 & 35.5 & 65.0 & 21.6 & 73.0 & \textbf{83.0} & 34.0 & \textbf{50.0} & 43.0 
    & 46.2 \\
    
    InternVL3-8B 
    & 40.0 & 39.0 & 10.0 & 71.5 & 44.5 & 30.8 & 65.0 & 25.2 & 77.0 & 82.0 & 36.0 & 21.7 & 50.0 
    & 45.6 \\
    
    InternVL3-78B 
    & 23.3 & 49.0 & 33.0 & 74.0 & 42.5 & 37.4 & 66.7 & 30.0 & 76.0 & 81.0 & 40.0 & 45.0 & 42.0 
    & 49.2 \\
    
    Qwen3-VL-8B 
    & 21.7 & 50.0 & 26.0 & \textbf{81.5} & 46.6 & 43.1 & 66.7 & 30.4 & 74.0 & 79.0 & 37.0 & 43.3 & 51.0 
    & 50.0 \\
    
    Qwen3-VL-235B-A22B 
    & \textbf{61.7} & \textbf{82.5} & 38.6 & 36.7 & 44.0 & 39.0 & 38.9 & \textbf{49.0} & 73.0 & 82.0 & 37.8 & 33.3 & 48.0 
    & 51.1 \\

    \multicolumn{15}{l}{\footnotesize \textit{\textcolor{darkgray}{Open-source Remote Sensing Vision-Language Models}}} \\
    GeoChat 
    & 16.7 & 29.0 & 2.0 & 23.0 & 21.1 & 16.8 & 35.0 & 24.2 & 33.0 & 43.0 & 10.0 & -- & 21.0 
    & 22.9 \\

    GeoLLaVA-8K 
    & 26.7 & 38.0 & \textbf{49.0} & 69.0 & 41.6 & 31.6 & 65.0 & 35.0 & 67.0 & 78.0 & \textbf{66.0} & \textbf{50.0} & \textbf{52.0} 
    & 51.5 \\
    
    DeepEyes 
    & 31.7 & 41.0 & 33.0 & 75.0 & 41.6 & 38.1 & \textbf{68.3} & 31.4 & 70.0 & 78.0 & 46.0 & \textbf{50.0} & 46.0 
    & 50.0 \\
    
    GeoEyes 
    & 38.3 & 40.0 & 24.0 & 73.5 & 59.5 & \textbf{66.1} & \textbf{68.3} & 32.2 & 77.0 & 80.0 & 56.0 & 40.0 & 50.0 
    & 54.2 \\

    ScaleEarth 
    & 41.7 & 52.3 & 44.8 & 78.6 & 56.2 & 53.4 & 66.1 & 39.7 & 75.2 & 81.6 & 57.3 & 49.1 & 50.2 
    & \textbf{57.4} \\    

    \bottomrule
    \end{tabular}
}
\vspace{1mm}
\begin{minipage}{0.98\textwidth}
\footnotesize
\end{minipage}
\end{table*}

\subsection{Comments on Evaluation Results}

The comparative results reveal that the relative advantages of RS-MLLMs and CV-MLLMs are highly task-dependent. On conventional remote sensing scene classification and VQA benchmarks, several RS-MLLMs achieve competitive or even superior performance, especially on datasets whose visual distributions are close to those used during RS-specific instruction tuning. For example, RS-oriented models show clear advantages on RSVQA-HR and visual grounding tasks, where domain-specific object categories, spatial expressions, and grounding annotations are important. However, this advantage becomes less consistent on recently proposed RS-MLLM-specific benchmarks, such as LHRS-Bench and FIT-RSRC. On these datasets, recent general-purpose CV-MLLMs, including GPT-4o, Qwen3-VL, GLM-4.6V-Flash, and Qwen2-VL, often outperform most existing RS-MLLMs. This suggests that current RS-MLLMs may still suffer from limited instruction diversity and insufficient generalization beyond their training distributions.

The progress of general-purpose CV-MLLMs is clearly reflected in their remote sensing performance. Compared with earlier models such as Qwen-VL-Chat, LLaVA-1.5, and MiniGPT-v2, recent models demonstrate substantially stronger visual perception, instruction following, and reasoning abilities. In particular, Qwen3-VL and GLM-4.6V-Flash achieve strong results on LHRS-Bench and FIT-RSRC, while InternVL3.5 also shows competitive performance on RSVQA-HR. These observations indicate that advances in general multimodal foundation models can transfer effectively to remote sensing tasks, even without explicit RS-specific fine-tuning.

The results also show that model size alone does not determine performance. Some relatively compact models, such as TinyLLaVA, outperform several RS-MLLMs on specific benchmarks, while larger or domain-tuned models may still fail on tasks requiring robust spatial reasoning, multi-image understanding, or fine-grained relation comprehension. This indicates that the quality and diversity of instruction data, the alignment strategy, and the reasoning capability of the base model are often more important than parameter scale alone.

Overall, these findings suggest that future RS-MLLM development should not rely solely on domain-specific supervised fine-tuning over narrow task formats. Instead, more attention should be paid to constructing diverse and carefully curated RS instruction datasets, covering multi-resolution imagery, multi-modal observations, multi-view inputs, multi-temporal change analysis, visual grounding, and compositional reasoning. In addition, stronger general-purpose MLLMs can serve as effective backbones or teachers for RS-MLLMs, while remote-sensing-specific training should focus on complementing their weaknesses in spatial reasoning, sensor-aware understanding, geographic knowledge, and precise localization. Such a combination of strong general multimodal capabilities and well-designed RS-specific adaptation is likely to be a more promising direction for building robust and practical RSISU systems.

\begin{figure*}[!tbh]
	\centering
	\includegraphics[width=0.96\textwidth]{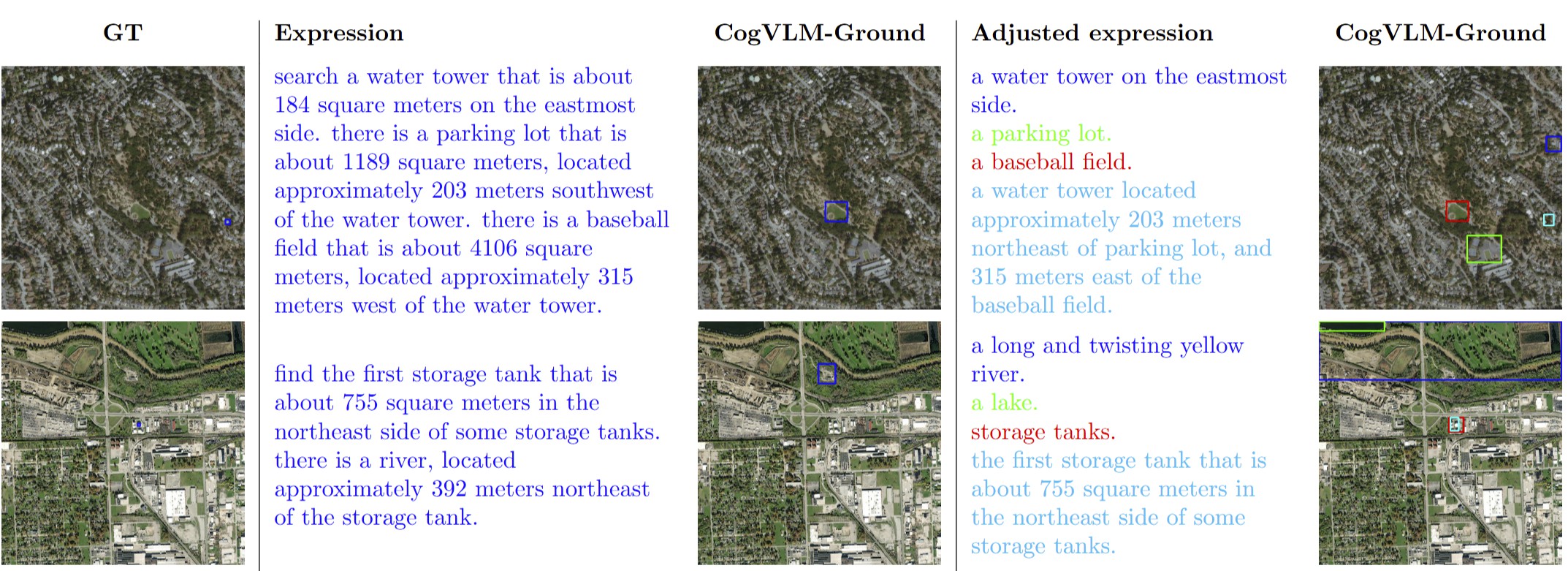}
\caption{Object-level verification improves remote-sensing visual grounding. The initial grounding result is inaccurate, while incorporating auxiliary object-level evidence refines the referring expression and leads to more accurate localization.}
	\label{fig:VG_wrong_correctProcess}
\end{figure*}

\begin{figure*}[!tbh]
	\centering
	\includegraphics[width=0.95\textwidth]{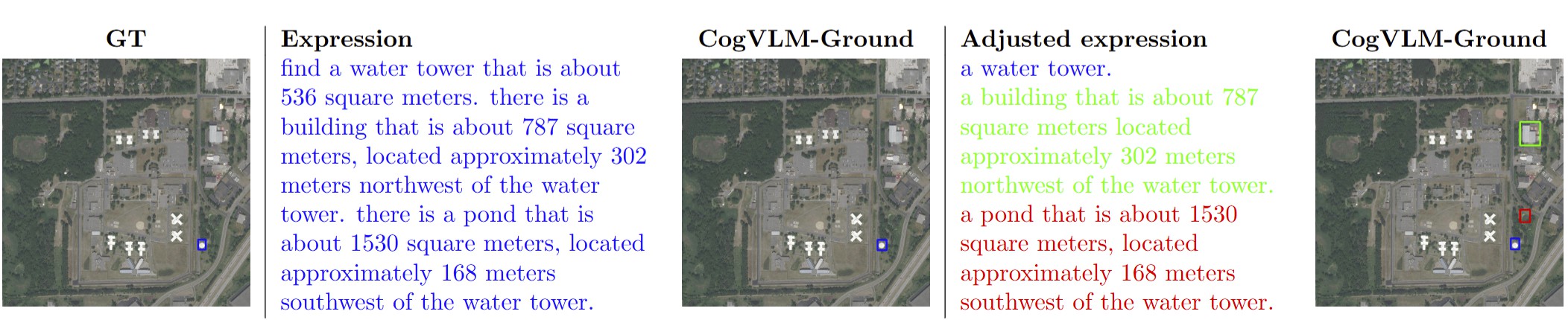}
\caption{Object-level verification reveals spurious visual grounding. Although the initial localization appears plausible, further examination of related objects exposes inconsistent object-level reasoning.}
	\label{fig:VG_correct_wrongProcess}
\end{figure*}

\section{Open Problems and Future Directions}
\label{sec:challenge}
Although RS-MLLMs have shown promising potential for remote sensing image understanding, the above comparisons reveal that their capabilities remain far from mature. Their performance is highly task-dependent: RS-MLLMs show clear advantages on visual grounding, RSVQA-HR, and XLRS-Bench. In contrast, recent general-purpose CV-MLLMs achieve stronger or comparable performance on LHRS-Bench, FIT-RSRC, and several scene classification settings. These findings indicate several important open problems for future research.

\begin{figure}[t]
\centering
\includegraphics[width=0.48\textwidth]{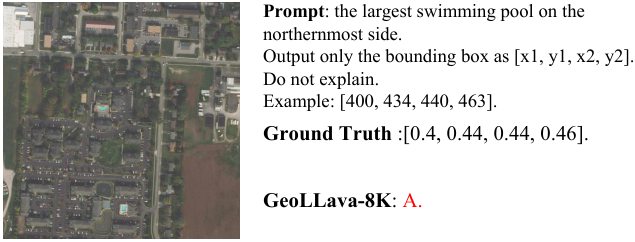}
\caption{Failure case of GeoLLaVA-8K due to the narrow instruction-tuning.}
\label{fig:failure}
\end{figure}

\subsection{Reliable and Contamination-aware Evaluation}
Fair and reproducible evaluation is increasingly important for RS-MLLMs. Existing results are drawn from heterogeneous sources, including literature reports, reproduced experiments, and API-based evaluations, which complicates direct comparison. As summarized in Table~\ref{tab:contamination_audit}, contamination risks vary across benchmarks and evaluated models. Future studies should therefore report benchmark and model release dates, potential training-data overlap, exact model or API versions, prompts, and evaluation protocols when presenting zero-shot results. Such transparency is essential for distinguishing zero-shot evaluation with a low risk of contamination from benchmark-aware inference and enabling reliable model comparisons.

\subsection{Generalization beyond Narrow RS Instruction Tuning}
Current RS-MLLMs do not consistently outperform general CV-MLLMs across all benchmarks. As shown in Fig.~\ref{fig:failure}, narrow instruction tuning limits the instruction-following and generalization capabilities of GeoLLaVA-8K. Although it performs well on ultra-high-resolution RS-VQA, it often produces invalid or inconsistent responses for out-of-distribution tasks such as scene classification and visual grounding, indicating limited adaptability to broader RSISU tasks. This suggests that some RS-specific models may overfit limited instruction formats, visual distributions, or task types. Future work should move beyond narrow supervised fine-tuning and construct more diverse RS instruction datasets. These datasets should cover different sensors, resolutions, geographic regions, object categories, question formats, and reasoning patterns, so that RS-MLLMs can generalize to unseen tasks rather than only performing well on familiar benchmarks.

\subsection{Spatial Reasoning and Fine-grained Grounding}
Spatial reasoning remains a major bottleneck for current MLLMs. The failure cases discussed above show that many models struggle with shadow-based sun-direction reasoning, left/right relations, object counting, orientation understanding, distance estimation, and precise visual grounding. These tasks require not only object recognition but also geometric and physical reasoning. Future RS-MLLMs should incorporate geometry-aware supervision, spatial relation verification, and region-level grounding ability to support reliable interpretation of complex remote sensing scenes.

\subsection{Improved Performance by Training-free Methodologies}

Developing RS-MLLMs usually requires substantial efforts in domain-specific data collection, annotation, and instruction tuning. However, our quantitative analysis shows that such resource investment does not necessarily lead to consistently superior performance. As discussed in Section~\ref{sec:performance}, several CV-MLLMs achieve competitive performance compared with RS-MLLMs on certain RS tasks, even without explicit adaptation to RS instruction data. This suggests that training-free methodologies, such as prompt engineering, in-context learning, reasoning-oriented prompting, and tool-assisted inference, remain promising but insufficiently explored directions for improving RS-MLLMs. Previous studies have shown that few-shot learning and Chain-of-Thought prompting can enhance the reasoning and generalization abilities of large models. In our experiments, as shown in Figure~\ref{fig:fs_comparison}, adding a one-shot example to the prompt increases the classification accuracy on EuroSAT from 56.7\% to 72.4\%, indicating the effectiveness of training-free adaptation for RS scene understanding.

We further observe that prompt formulation can substantially affect model behavior, particularly for tasks requiring fine-grained grounding or structured outputs. For example, the prompt ``\textless phrase\textgreater A chimney on the right.\textless /phrase\textgreater'' enables KOSMOS-2 to produce accurate referring expression comprehension results, whereas the plain prompt ``A chimney on the right.'' fails to activate the desired grounding capability. Minor formatting changes can also influence the output; for instance, removing the period in ``\textless phrase\textgreater A chimney on the right\textless /phrase\textgreater'' leads to inferior results. These observations indicate that prompt design is an important factor affecting the reliability of MLLMs in RS applications, especially in multi-class classification and visual grounding tasks, where task instructions, output formats, and grounding markers directly influence predictions, as illustrated in Figure~\ref{fig:VG_wrong_correctProcess} and Figure~\ref{fig:VG_correct_wrongProcess}. Nevertheless, the effectiveness of these strategies is model- and task-dependent. Therefore, future RS-MLLM studies should systematically evaluate task-aware and model-specific training-free strategies before relying on costly domain-specific instruction tuning.
\begin{table}[!t]  
  \centering   
  \caption{RSISU tasks supported by model-tool collaboration. Red fonts denote questionable results.}  
  \label{tab:rs-agent}  
  \includegraphics[width=0.48\textwidth]{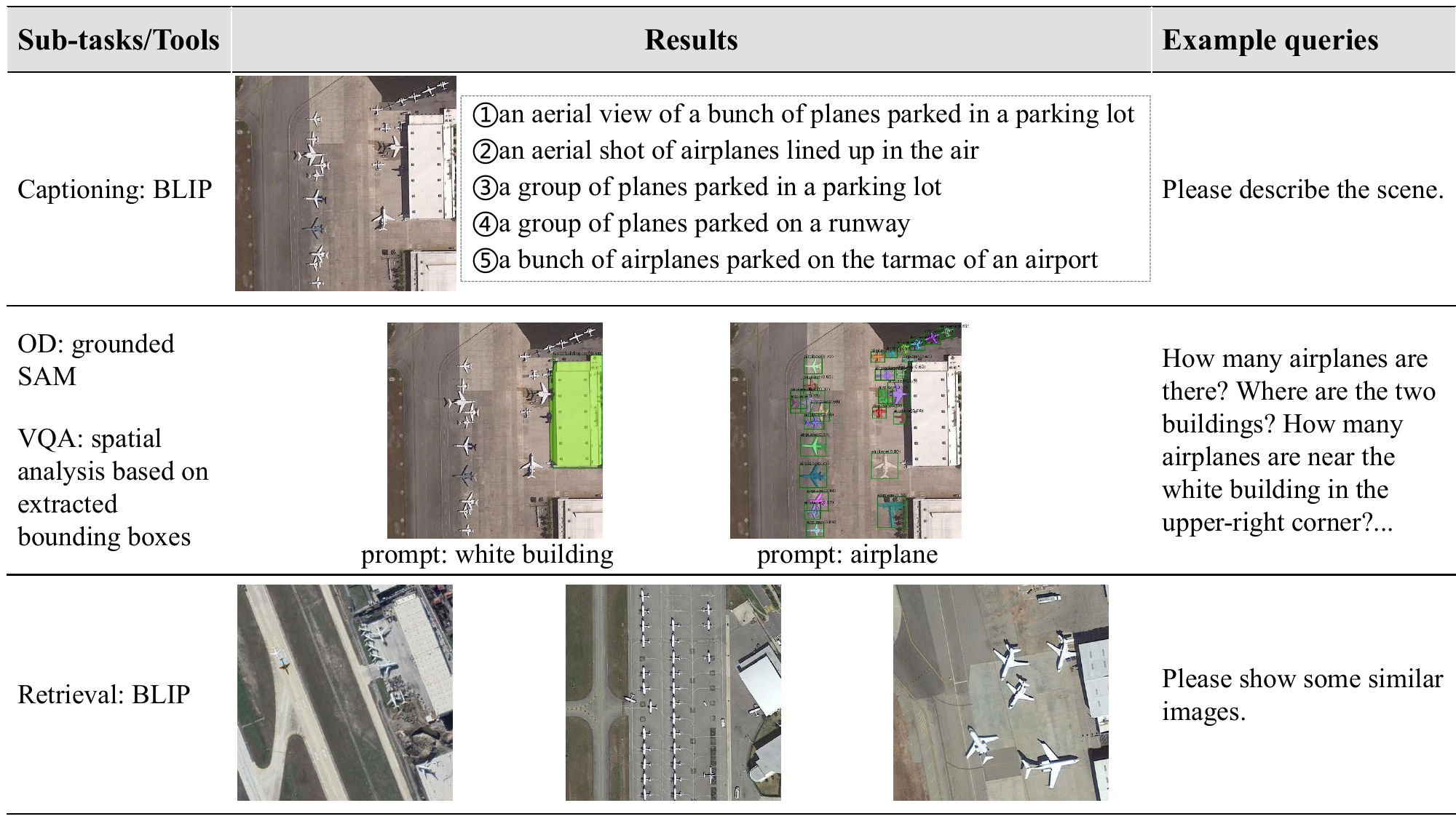}  
\end{table} 

\subsection{Multi-modal, Multi-image, and Ultra-high-resolution Understanding}

Real-world RSISU often involves heterogeneous inputs, including true-color optical images, false-color composites, SAR, infrared, multispectral, hyperspectral, multi-view images, and multi-temporal observations. Although recent efforts such as Earth-OneVision~\citep{earth-onevision} and RemoteShield~\citep{remoteshield} have explored multimodal MLLMs for remote sensing, most existing RS-MLLMs still rely primarily on single-modality inputs, particularly RGB imagery, leaving multi-source RS data insufficiently exploited. In addition, ultra-high-resolution RS images contain rich local details but are difficult to process due to resolution and context-length limitations. Future models should support sensor-aware encoding, multi-image comparison, temporal reasoning, adaptive region selection, and hierarchical high-resolution understanding.

\subsection{Efficient and Tool-augmented RSISU Systems}
Building and deploying RS-MLLMs require substantial computational resources, especially for high-resolution inputs and large-scale instruction tuning. Therefore, efficiency should be a key future direction, including lightweight architectures, visual token compression, parameter-efficient tuning, and training-free adaptation methods such as prompt engineering and few-shot prompting. In addition, RS agents that combine MLLMs with specialized tools, such as object detectors, segmentation models, retrieval systems, and GIS software, offer a practical path toward reliable RSISU. As shown in Table~\ref{tab:rs-agent}, integrating BLIP~\citep{Li2022BLIPBL} enables more comprehensive image caption generation, while SAM~\citep{Kirillov2023SegmentA} provides fine-grained spatial information that facilitates grounding queries in MLLMs. In addition, the directions discussed for RS agents in Sec~\ref{subsection:rs_agent}, such as reliable tool selection, long-horizon reasoning, trajectory verification, multi-agent collaboration, and domain-specific geospatial workflows, remain important topics for further exploration.

\section{Conclusions}
\label{sec:conclusion}

This survey provides a systematic review and diagnostic evaluation of MLLMs for remote sensing image scene understanding. Beyond summarizing model architectures, training data, and adaptation strategies, our analysis supports four main conclusions about the current development of RS-MLLMs.

First, the observed advantages of RS-MLLMs are localized rather than universal. RS-oriented models perform strongly on several domain-specific tasks, particularly remote sensing VQA, visual grounding, and ultra-high-resolution image understanding. These advantages are most evident when the evaluation data and task formats are closely related to those used for RS instruction tuning. However, they do not yet translate into consistent superiority across diverse RSISU tasks. Current RS-specific adaptation therefore improves selected capabilities but has not established general remote sensing understanding.

Second, general foundation capabilities increasingly determine benchmark performance. Recent general-purpose CV-MLLMs achieve competitive or superior results on several RS benchmarks without explicit RS-specific fine-tuning. Their performance is supported by strong visual perception, instruction following, and compositional reasoning. These capabilities can be more influential than model size or narrow domain-specific training. General-purpose MLLMs should therefore be considered important backbones or teachers for future RS-MLLMs rather than merely external baselines.

Third, the evaluation protocol has become a central bottleneck. Existing comparisons often combine literature-reported results, reproduced experiments, API-based evaluations, and transfer-learning results. Potential benchmark contamination further complicates their interpretation. Future studies should clearly distinguish strict contamination-free zero-shot evaluation, no-adaptation inference, benchmark-aware evaluation, and transfer learning. They should also report model versions, benchmark release dates, training-data overlap, prompts, and output-processing rules. Without such controls, the benefits of RS-specific adaptation may be overestimated.

Finally, the next frontier lies in sensor-aware and verifiable remote sensing intelligence. Future research should move beyond repeatedly expanding conventional captioning and VQA datasets. Greater attention is needed for multisensor, multi-image, and multitemporal understanding. Models must also improve geometric reasoning, precise grounding, geographic knowledge, and robustness across task formats. Tool use, external knowledge, and agent-based verification provide promising ways to make model outputs more traceable and reliable.

The central conclusion is therefore clear: domain specificity alone is insufficient for building general and practical RS intelligence. Future progress will depend on combining strong general-purpose multimodal foundations with targeted RS adaptation. Such adaptation should complement the weaknesses of general models through sensor-aware perception, spatially grounded reasoning, and geographic knowledge.

\section*{Acknowledgments}

This work was supported in part by the National Natural Science Foundation of China under Grant 62525108, Grant 62371185, Grant 62006241, and Grant 42201513, in part by the National Key Research and Development Program of China under Grant 2021 YFA0715203, in part by the Science and Technology Innovation Program of Hunan Province under Grant 2024RC1030 and Grant 2023RC3124, in part by the Project of Yuelushan Center for Industrial Innovation under Grant 2025YCII0202, in part by the China Postdoctoral Science Foundation under Grant 2022M723902 and Grant 2023T160789.

\section*{Declaration of Interest Statement}
The authors declare no competing financial interests or personal relationships that could have influenced the work reported in this paper.

\bibliographystyle{elsarticle-harv}
\biboptions{authoryear}

\bibliography{reference}

\end{document}